%% file: ACL2025.tex
\PassOptionsToPackage{dvipsnames, table}{xcolor}
\documentclass[11pt]{article}
\RequirePackage[dvipsnames, table]{xcolor}  

\usepackage{acl}
\usepackage{enumitem}
\usepackage{graphicx}
\usepackage{subcaption} 
\usepackage{makecell}
\usepackage{amsmath}
\usepackage{graphicx}
\usepackage{calc}
\usepackage{footmisc}
\usepackage{amsmath}
\usepackage{booktabs}
\usepackage{siunitx}
\usepackage{enumerate}  
\usepackage{multirow}
\usepackage{float} 
\usepackage{acl}
\usepackage{placeins}
\usepackage{times}
\usepackage{latexsym}
\usepackage{array}
\usepackage{adjustbox}
\usepackage[T1]{fontenc}

\usepackage[utf8]{inputenc}

\makeatletter
\input{tcolorbox.sty}
\makeatother

\usepackage{fancyvrb}
\usepackage{caption}
\usepackage{newfloat}
\usepackage{fvextra}
\usepackage{amsmath}

\DeclareFloatingEnvironment{boxes} 

\newtcolorbox{theoremblock}[1][]{colframe=white, colbacktitle=green!80!black,
  colback=green!10, coltitle=white, boxrule=0pt, rounded corners=all, 
  fonttitle=\bfseries, title=#1}

\newtcolorbox{propositionblock}[1][]{colframe=white, colbacktitle=red!80!black,
  colback=red!10, coltitle=white, boxrule=0pt, rounded corners=all, width=\textwidth,
  fonttitle=\bfseries, title=#1}

\newtcolorbox{definitionblock}[1][]{colframe=white, colbacktitle=blue!80!black,
  colback=blue!10, coltitle=white, boxrule=0pt, rounded corners=all, width=\textwidth,
  fonttitle=\bfseries, title=#1}

\newtcolorbox[number within=section]{lemmablock}[1][]{
colframe=white, colbacktitle=blue!50!black,
  colback=blue!5, coltitle=white, boxrule=0pt, rounded corners=all, width=\textwidth,
  fonttitle=\bfseries, title=#1,  left=0mm, right=0mm}

\DefineShortVerb{\|} 

\usepackage{microtype}

\usepackage{inconsolata}

\title{\includegraphics[height=1.4em]{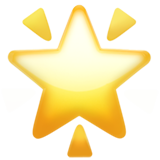}ASTRA: A Negotiation Agent with Adaptive and Strategic Reasoning via Tool-integrated Action for Dynamic Offer Optimization
}

\author{Deuksin Kwon$^{1,4}$\hspace{0.3cm}Jiwon Hae$^1$\hspace{0.3cm}Emma Clift$^{2}$\hspace{0.3cm}Daniel Shamsoddini$^{3}$\hspace{0.3cm}\\\textbf{Jonathan Gratch}$^{1,4}$\hspace{0.3cm}\textbf{Gale M. Lucas}$^{1,4}$\\
$^1$University of Southern California
$^2$St. Olaf College
$^3$Northwestern University\\
$^4$USC for Institute of Creative Technologies\\
\texttt{\{deuksink, hae\}@usc.edu}
\\\texttt{emma.m.clift@gmail.com}
\texttt{dshamsoddini@u.northwestern.edu}
\\\texttt{\{gratch, lucas\}@ict.usc.edu}}

\begin{document}
\maketitle

\begin{abstract}
\input{Section/0_abs}  
\end{abstract}

\input{Section/1_intro}
\input{Section/3_method}

\input{Section/4_experiment}
\input{Section/5_results}
\input{Section/6_discussion}
\input{Section/2_related_works}

\input{Section/7_conclusion}

\clearpage
\input{Section/8_limitation}

\input{Section/99_acknowledgment}
\bibliography{anthology,custom}
\clearpage
\appendix
\input{Section/9_appendix}

\end{document}

%% file: Section/0_abs.tex
Negotiation requires dynamically balancing self-interest and cooperation within the flow of conversation to maximize one’s own utility. Yet, existing agents struggle due to bounded rationality in human data, low adaptability to counterpart behavior, and limited strategic reasoning. To address this, we introduce principle-driven negotiation agents, powered by ASTRA, a novel framework for turn-level offer optimization grounded in two core principles: opponent modeling and Tit-for-Tat reciprocity. ASTRA operates in three stages: (1) interpreting counterpart behavior, (2) optimizing counteroffers via a tool-integrated action with a linear programming (LP) solver, and (3) selecting offers based on strategy assessment and the partner’s acceptance probability. Through simulations and human evaluations, our agent effectively adapts to an opponent’s shifting stance and achieves favorable outcomes through enhanced adaptability and strategic reasoning. Beyond enhancing negotiation performance, it also serves as a powerful coaching tool, offering interpretable strategic feedback and optimal offer recommendations beyond human bounded rationality, with its potential further validated through human evaluation. Code is available at \url{https://github.com/DSincerity/ASTRA-NegoAgent}.

%% file: Section/1_intro.tex
\section{Introduction}
\label{sec:intro}


Negotiation is an inherently dynamic conversational process that requires balancing self-interest with cooperation to maximize personal gains through mutual agreement~\cite{chawla-etal-2023-selfish}. At its core lies the strategic navigation of dialogue to identify the optimal profit point (OPP), the best deal acceptable to the counterpart. Reaching the OPP requires more than proposing fair offers; it demands continuous interpretation of linguistic cues and responses with strategic counteroffers~\cite{olekalns2008emergent}. As illustrated in Figure 1, this dynamic adaptation hinges on interpreting linguistic signals indicating the counterpart’s \textit{fairness} (i.e., the degree to which an offer distributes points evenly) and \textit{stance} (i.e., a shift in generosity or greed from the previous offer, or remaining neutral). Successful negotiation thus relies on adapting to evolving behavioral signals expressed through language in conversations~\cite{heunis2024strategic, brzostowski2006adaptive}.
\begin{figure}[t]
    \centering
    \includegraphics[width=1\linewidth]{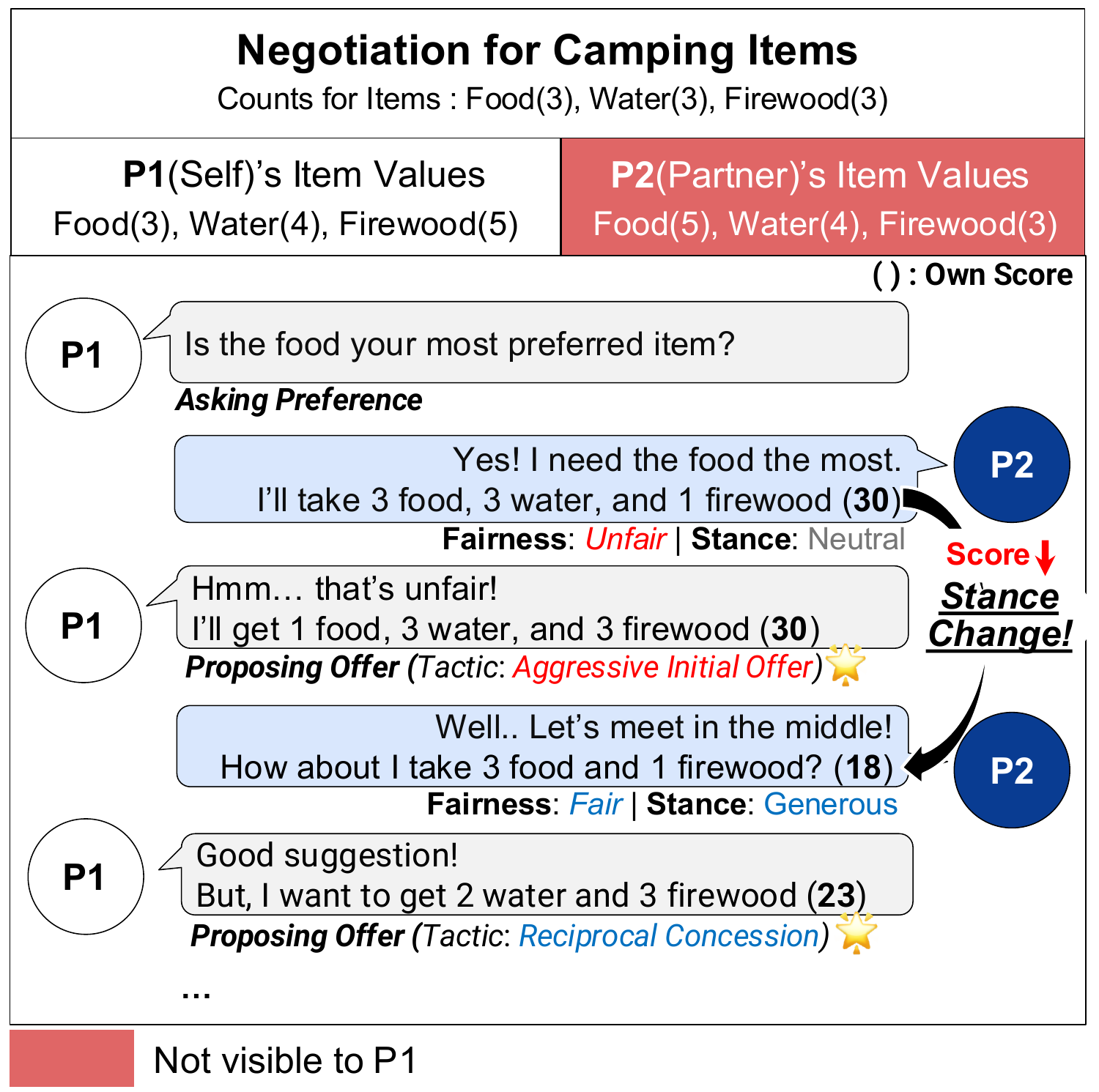}
    \caption{Example negotiation dialogue showcasing the strategic dynamics between players. P1 effectively guides the negotiation by capturing the partner’s (P2) shifting behavioral signals (e.g., \textit{fairness}, \textit{stance}) and generating strategic responses based on diverse tactics.}
    \label{fig:intro}
\end{figure}


Existing negotiation agents fall broadly into two categories. The first includes numerical offer-based agents that maximize outcomes without using natural language~\cite{jonker2017automated}, making them poorly suited for real-world negotiations. The second involves dialogue-based agents trained on human-to-human data via supervised learning and self-play Reinforcement Learning (RL)~\cite{lewis-etal-2017-deal, kwon2023reward, yang-etal-2021-improving, zhang-etal-2020-learning}. However, these agents inherit limitations from human data, including bounded rationality due to cognitive constraints and emotions, which limit strategic flexibility~\cite{simon1990bounded}. Although RL partially mitigates this, reliance on static and final-score rewards promotes rigid negotiation styles, hindering the agent’s ability to dynamically adapt and reach the OPP. These existing approaches also lack explicit use of established negotiation principles grounded on observed opponent behaviors, which is crucial for successful negotiation outcomes~\cite{lewis2017deal, chawla-etal-2023-selfish}.

Moreover, while LLMs have shown strong reasoning capabilities and promise in negotiation dialogues, their effectiveness has been largely confined to simpler, single-issue bargaining~\cite{shea2024ace, bianchi2024well}. Notably, they struggle significantly in multi-issue bargaining (MIB)~\cite{fershtman1990importance}, a widely used framework in negotiation research that captures the complexity and diversity of real-world negotiations involving multi-dimensional trade-offs~\cite{zhang2024llm, schneider2023negotiating, kwon-etal-2024-llms}. This reveals a critical gap and underscores the need to enhance their strategic reasoning by incorporating principles such as Tit-for-Tat (TFT) reciprocity (conditional cooperation) and opponent modeling, as well as improving their capability to generate optimal offers.

To address these challenges, we propose a principle-driven LLM-based negotiation dialogue system powered by \textit{ASTRA} (\textbf{A}daptive and \textbf{ST}rategic \textbf{R}easoning via a tool-integrated \textbf{A}ction), a novel framework for turn-level offer decision through dynamic adaptation. \textit{ASTRA} is grounded in established negotiation principles—opponent modeling and TFT reciprocity~\cite{kelley1996classroom, baarslag2016learning}—and strategically leverages LLMs at key stages of the process: (1) the opponent modeling stage, which identifies the counterpart’s behavioral signals through linguistic cues, (2) the offer optimization stage, which generates optimal offers using a linear programming (LP) solver guided by a dynamic reward mechanism, and (3) the final offer decision stage, which reasons over multiple negotiation tactics and partner's accepctance probability with a virtual partner agent to select the best offer. The system also incorporates LLM-based modules for priority inquiry and inconsistency detection, further enhancing its opponent modeling and overall strategic adaptability.

Simulations with diverse agent types and human evaluations confirm that \textit{ASTRA} dynamically adapts to an opponent’s shifting stance, whether greedier or more generous, balancing cooperation and self-interest under TFT to effectively achieve the OPP. Beyond enhancing negotiation agents, ASTRA functions as a powerful coaching tool, offering explainable, explicit strategic feedback~\cite{johnson2019using} and optimal offer suggestions, validated through human evaluation confirming its ability to generate novel, advantageous offers beyond bounded rationality. The contributions of our work are as follows:


\begin{itemize}
\setlength{\itemsep}{0pt}
\item We introduce \textit{ASTRA}, a principle-driven LLM-based framework for turn-level strategic reasoning and offer generation in negotiation dialogue systems, validated through both large-scale simulations and human evaluation.  
\item Our framework achieves expert-level strategic reasoning without additional training, overcoming the limits of small datasets and non-expert demonstrations.  
\item By integrating LP-based actions, \textit{ASTRA} ensures numerical consistency, enables context-aware offer optimization, and enhances decision reliability in negotiation.  
\item Beyond advancing negotiation agents, \textit{ASTRA} serves as a coaching and decision-support tool, providing interpretable feedback and real-time strategic insights, with human evaluation confirming its ability to generate novel, advantageous offers beyond bounded rationality.  
\end{itemize}



%% file: Section/3_method.tex
\section{Methods}

\subsection{Components for Opponent Modeling}

Opponent modeling is a crucial principle for effective negotiation, enabling agents to infer a counterpart’s unknown preferences and strategies~\cite{kelley1996classroom, chawla-etal-2022-opponent}. At each turn, understanding the partner’s offers and inferring their preferences is key to optimizing counteroffers. To achieve this, we integrate three core components into our negotiation agent:

\begin{itemize}
\setlength{\itemsep}{0pt}
\item Preference Asker: Strategically inquires about the partner’s preferences early on.
\item Preference Consistency Checker: Ensures inferred partner preferences (\textit{IPP}) align with observed behavior (see Appendix \ref{append:CC}).
\item	Preference Updater: Refines \textit{IPP} dynamically based on new partner utterances.
\end{itemize}
End-to-end models often struggle with complex negotiations, lacking clarity and control. Our modular approach enhances adaptability, interpretability, and future scalability.

\subsection{Response Modes}\label{res_mode}

\begin{figure*}[htb!]
  \centering
  \includegraphics[width=0.97\textwidth]{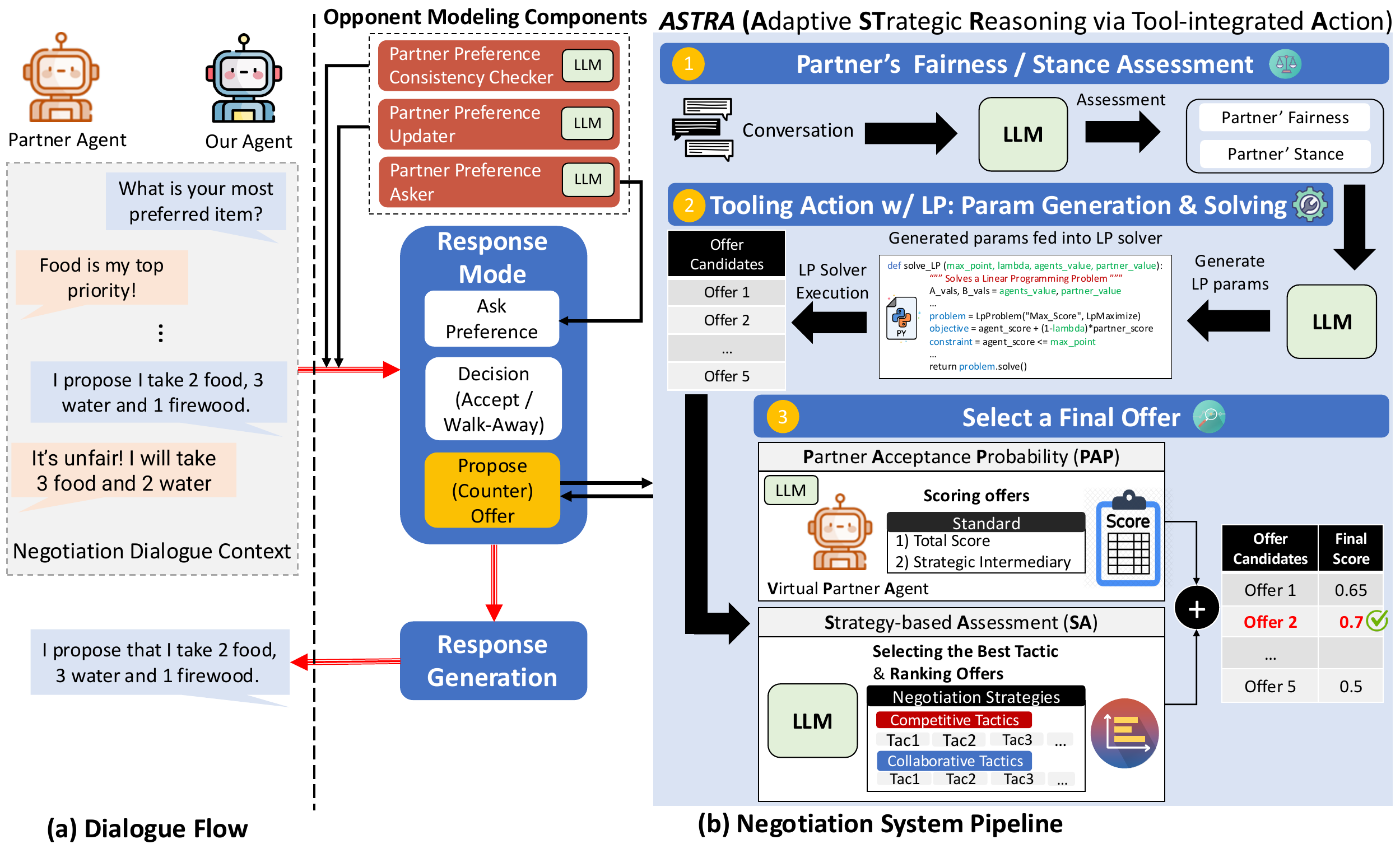}
  \caption{An illustration of our agent’s adaptive strategic reasoning process and response generation. (a) depicts the dialog flow, while (b) presents the response generation pipeline with internal modules and the~\textit{ASTRA}. }
  \label{fig:main} 
\end{figure*}


At each turn, the agent selects one of three response modes as shown in the Figure~\ref{fig:main}:
\begin{itemize}
\setlength{\itemsep}{0pt}
\item Asking Preference: Early in the negotiation, if no \textit{IPP} exist, the agent asks about preferences. If the response is unclear, it assumes preferences opposite to its own.
\item Decision on Partner’s Offer: The agent accepts an offer if its score meets or exceeds its most recent offer. It walks away if the partner (a) makes no concessions for three consecutive turns or (b) twice offers below the agent’s Best Alternative to a Negotiated Agreement (BATNA) despite a walk-away warning. These decisions balance short-term gains with long-term bargaining power, reputation, and future negotiation prospects.
\item Proposing Offer: When \textit{IPP} exists and neither of the above response modes apply, the agent uses the \textit{ASTRA} framework to generate adaptive and strategic offers.
\end{itemize}

\subsection{Offer Optimization with \textit{ASTRA}}

When making an offer or counteroffer, the agent employs the three-stage \textit{ASTRA} framework to adaptively and strategically respond to the opponent’s behavior. It is grounded in TFT, a proven reciprocity-based principle that mirrors the opponent’s actions \cite{axelrod1981evolution, baarslag2013tit}. \textit{ASTRA} consists of three key stages: (1) \textit{fairness} and \textit{stance} assessment, (2) offer optimization via linear programming (LP), and (3) final offer selection with strategy and partner acceptance probability assessment. These stages enable the agent to maximize utility while remaining strategically adaptive.

\paragraph{Stage 1: Opponent Modeling—Fairness and Stance Assessment} 
The perception of fairness is crucial in negotiation, as individuals often equate fairness with equitable resource distribution or similar utility levels \cite{welsh2003perceptions}. A TFT principle requires understanding the partner’s stance, which the first stage achieves by assessing their \textit{fairness} and \textit{stance} based on prior offers, providing a foundation for the agent’s strategic adaptability. \textit{Fairness} is determined by the point gap between the agent’s score (\( S_{\text{a}}^{P} \)) and the partner’s score (\( S_{\text{p}}^{P} \)) in the partner’s offer (P), as well as the absolute level of \( S_{\text{p}}^{P} \) (e.g., half of the possible max score (PMS)); see Appendix~\ref{append:first_astra} for details on the fairness threshold \( \theta_f \):

\vspace{-1.5ex}
\begin{equation}
\small
\text{Fairness} =
\begin{cases} 
\text{Fair}, & \text{if } \lvert S_{\text{a}}^{\text{P}} - S_{\text{p}}^{\text{P}} \rvert \leq \theta_f \text{ or } S_{\text{p}}^{\text{P}} \leq \frac{\text{PMS}}{2}  \\
\text{Unfair}, & \text{otherwise}.
\end{cases}
\label{eq:fairness}
\end{equation}

The partner's stance is determined by the change between their scores ($S_{\text{p}}^{\text{P}, (t)}$ and $S_{\text{p}}^{\text{P}, (t-1)}$) in two most recent offers made by the partner at time $t$ :

\begin{equation*}
\small
\Delta S_{\text{p}} = S_{\text{p}}^{\text{P}, (t)} - S_{\text{p}}^{\text{P}, (t-1)} 
\end{equation*}
\vspace{-2ex}
\begin{equation}
\small
\text{Stance} =
\begin{cases} 
\text{Generous} & \text{if } \Delta S_{\text{p}} < 0, \\
\text{Neutral} & \text{if } \Delta S_{\text{p}} = 0, \\
\text{Greedy} & \text{if } \Delta S_{\text{p}} > 0.
\end{cases}
\end{equation}

\paragraph{Stage 2: Tooling Action with Linear Programming for Offer Optimization}


At this stage, the agent uses an LP solver as a tool to generate optimal offers, dynamically adjusting parameters in line with the TFT principle. It formulates negotiation as an optimization problem, sets the LP objective, and executes the solver via a Python function, incorporating intermediate results into its decisions. The agent modulates two key parameters: $\lambda$, which balances its own score ($S_{\text{a}}$) and the partner’s score ($S_{\text{p}}$), and $S_{\text{max}}$, the upper bound on its score. $\lambda$ ($0 \leq \lambda \leq 1$) controls the agent’s stance, with 1 favoring self-interest and 0 indicating cooperation. Implementation details are in Figure~\ref{appendix:lp_codes} (Appendix). The LP objective is defined as follows:

\vspace{-1ex}
\begin{equation*}
\text{Objective}:
\text{max} \: S_{\text{a}} + (1 - \lambda) * S_{\text{p}}
\label{eq:lp_objective}
\end{equation*}
\vspace{-4ex}
\begin{equation}
\text{subject to:} \: 
S_{\text{a}} \leq S_{\text{max}}
\label{eq:lp_constraint}
\end{equation}

To reduce the risk of suboptimal offers from inaccurate \textit{IPP} and to explore multiple optimal options, the agent solves the LP multiple times with varied parameters. Specifically, $\lambda$ is sampled within $\pm 0.3$ of the agent-determined value in 0.1 increments, and $S_{max}$ is explored in 1-point decrements down to 10 points lower. From these runs, the agent selects the top N offers by score, generating a set of adaptive and optimal candidates.

\paragraph{Stage 3: Final Strategic Offer Selection}\label{astra_third}

The third stage of \textit{ASTRA} involves the strategic selection of a final offer from the candidate set in Stage 2. Each offer is evaluated using two components: (1) Partner’s Acceptance Probability (\textit{PAP}) and (2) Strategy-based Assessment (\textit{SA}).

\textit{PAP} estimates the likelihood of partner acceptance. A Virtual Partner Agent (VPA) scores each offer based on the partner’s Total Score (\textit{TS}) and its effectiveness as a Strategic Intermediary (\textit{SI}) toward the ideal outcome. Each offer is evaluated five times, with average \textit{TS} and \textit{SI} normalized to a 10-point scale. Giving priority to \textit{TS}, \textit{PAP} is computed as \( \textit{PAP} = w \cdot TS + (1 - w) \cdot SI \), where \( w \) is a weighting parameter (see Appendix~\ref{append:trd_astra}).


\textit{SA} evaluates each offer based on alignment with nine turn-level negotiation tactics, grouped into two strategies: \textit{Competitive} (four focused on self-gain) and \textit{Collaborative} (five promoting mutual benefit). Tactic details are in Table~\ref{append:turn=level-offer_desc} in the Appendix. The agent selects the best-fit tactic, ranks offers, and repeats this process five times for robustness. The majority tactic is chosen, and a final min-max Normalized Ranking Score is computed.

The optimal offer is selected by combining \textit{PAP} and \textit{SA} into a weighted sum:

\vspace{-1ex}
\begin{equation}
\text{Final Score} = \alpha \cdot \textit{PAP} + \beta \cdot \textit{SA}.
\label{eq:pap_sa}
\end{equation}
Here, $\alpha$ and $\beta$ are empirically determined weights. The offer with the highest final score is selected and presented in the agent’s response, demonstrating its ability to generate adaptive and strategic offers.

%% file: Section/4_experiment.tex




\begin{table*}[thb]
\centering
\renewcommand{\arraystretch}{1}
\resizebox{1\linewidth}{!}{%
\begin{tabular}{ll|cccc}
\hline
\multicolumn{2}{c|}{Players}   & \multicolumn{4}{c}{Evaluation Metric}   \\ \hline
\multicolumn{1}{c}{\begin{tabular}[c]{@{}c@{}}Player-1\\ (P1)\end{tabular}} & \multicolumn{1}{c|}{\begin{tabular}[c]{@{}c@{}}Player-2\\ (P2)\end{tabular}} & \begin{tabular}[c]{@{}c@{}}Avg. Score (All)\\ (P1 vs P2)\end{tabular} & \begin{tabular}[c]{@{}c@{}}Avg. Score (Agreement)\\ (P1 vs P2)\end{tabular} & \multicolumn{1}{l}{\begin{tabular}[c]{@{}l@{}}T-statstic\\ (Agreement)\end{tabular}} & Walk-Away (\%) \\ \hline
\multicolumn{1}{l|}{Ours}   &  & {\underline{20.69} vs 12.33}   & \underline{24.34} vs 14.50   & \multicolumn{1}{l}{}  & 17\\
\multicolumn{1}{l|}{Ours wo/ ASTRA} &  & \underline{19.92} vs 16.16 & \underline{20.75} vs 16.83   & 8.55* & 4 \\
\multicolumn{1}{l|}{RL Agent~\cite{chawla-etal-2023-selfish}}   &  & 18.86 vs 19.06  & 18.86 vs 19.06& 9.49* & 0 \\
\multicolumn{1}{l|}{ICL-AIF~\cite{fu2023improving}}&  & 17.39 vs 16.65  & 18.22 vs 17.75& 13.27*& 9 \\
\multicolumn{1}{l|}{Pro-CoT~\cite{deng-etal-2023-prompting}} & \multirow{-5}{*}{Partner-Base}   & 12.45 vs 12.53  & 17.84 vs 17.98& 8.24* & 42\\ \hline
\multicolumn{1}{l|}{Ours}   & ICL-AIF  & \underline{17.82} vs 12.28  & \underline{21.95} vs 15.2   &   & 22\\
\multicolumn{1}{l|}{Ours}   & Pro-CoT   & \underline{14.52} vs 10.14 & \underline{22.31} vs 15.17   &   & 31\\
\multicolumn{1}{l|}{Ours}   & Partner-Greedy   & 3.65 vs 2.43 & \underline{24.33} vs 16.20   & \multicolumn{1}{l}{}  & {\color[HTML]{000000} 85} \\
\multicolumn{1}{l|}{Ours}   & Partner-Fair & \underline{16.41} vs 9.05  & \underline{25.24} vs   13.92 & \multicolumn{1}{l}{}  & {\color[HTML]{000000} 35} \\ \hline
\end{tabular}}
\caption{ Agent-to-agent simulation results. "Ours" denotes our negotiation agent powered by \textit{ASTRA}. “All” includes all negotiation cases, while “Agreement” includes only cases where an agreement was reached. Underlined scores in the Avg. Score columns indicate a statistically significant advantage over the partner (P2), based on a paired T-test (p < 0.001). Asterisks (*) indicate a significant score difference based on a T-test (p < 0.001) between Ours and each baseline when paired with the same partner.}
\label{tab:main_results}
\vspace{-2ex}
\end{table*}


\section{Experiment}

\subsection{Dataset} Our method enables an LLM-based agent to operate without a specific training dataset. However, our negotiation scenarios are based on the CaSiNo dataset~\cite{chawla2021casino}, involving two participants negotiating as campsite neighbors over packages of food, water, and firewood with individual item preferences. The CaSiNo dataset is also used to train the RL baseline; an example dialogue and details are provided in Appendix~\ref{append:casino_detail}.

\subsection{Agents Simulation} To evaluate the effectiveness of \textit{ASTRA}, we conduct agent-to-agent simulations following prior self-play and dyadic simulation studies~\cite{abdelnabi2024cooperation, lewis-etal-2017-deal}. We design diverse agent types, such as greedy and fair agents, using both prompting and RL methods~\cite{chawla-etal-2023-selfish, kwon2023reward}. This approach rigorously evaluates \textit{ASTRA}’s robustness and adaptability across diverse negotiation dynamics. For the opponent modeling component of our agent and \textit{ASTRA}’s decision-making, we use GPT-4o-mini (\texttt{2024-07-18}) and GPT-4o (\texttt{2024-08-06}), respectively. The partner agent is primarily powered by GPT-4o, and each experimental case in Table~\ref{tab:main_results} is simulated 100 times. In addition, simulations with other LLMs, including Gemini-2.0-Flash (\texttt{2025-02}) and Claude-3.5-Sonnet (\texttt{2024-06-20}) as partner agents, are conducted 50 times each. Decoding parameters of LLMs for the simulations are listed in Appendix~\ref{append:hyperparameters}. All simulation codes are publicly available\footnote{\url{https://github.com/DSincerity/ASTRA-NegoAgent}}.

For the baseline, we adopt three negotiation agents with strong performance in the negotiation: (1) the RL agent from \citet{chawla-etal-2023-selfish}, which uses a two-stage RL approach with a tailored reward function; (2) \textit{Pro-CoT}~\cite{deng-etal-2023-prompting}, which leverages Chain-of-Thought prompting to strategically and proactively guide negotiations; and (3) \textit{ICL-AIF}~\cite{fu2023improving}, which iteratively enhances its performance through AI feedback on negotiation strategies. We compare their outcomes with those of our \textit{ASTRA}-powered agent.


As objective metrics for simulations, we assess both individual and joint points and track walk-away cases, as both agreements and strategic walk-aways are key to effective negotiation.

\subsection{Human Evaluations}
To complement simulation outputs, we conduct human evaluations using a single subjective metric, \textit{Strategicness}, which assesses how well an offer aligns with the agent’s goal of maximizing negotiation outcomes.

Each evaluation instance includes an agent response and its dialogue history. We sample 50 cases from simulation logs, balancing context length and negotiation type; integrative (players have opposing preference) and distributive (players have the same preference). For each instance, two Partner Agent responses are generated from the same context. Each instance is then randomly assigned to one of two mutually exclusive test sets following a randomized design. 

Six expert annotators, knowlegeable in multi-issue bargaining, independently evaluate 50 cases each, with three annotators per set. They rate \textit{Strategicness} on a four-point scale, based on how effectively each response advances the agent’s negotiation goal. An example is shown in Figure~\ref{append:human eval}.

%% file: Section/5_results.tex
\section{Results}

\subsection{Effect of \textit{ASTRA}}

To evaluate \textit{ASTRA}, we ran simulations comparing scenarios with and without it. As shown in Table~\ref{tab:main_results}, walk-away rates differed significantly. All walk-away cases stemmed from deadlocks where the partner agent refused concessions, prompting our agent to walk away strategically.

Effectiveness was evaluated using agreement cases, excluding walkaways. Agent scores, computed as the inner product of obtained items and preference values, reflect negotiation success. Agents with \textit{ASTRA} averaged 24.31 points, significantly outperforming those without it (20.75; t = 8.55, p < 0.001), suggesting that \textit{ASTRA} helps secure an additional middle-priority item. All baselines, including the RL agent, achieved fair deals but underperformed in utility maximization, showing statistically significant score differences from our agent (t-test, p < 0.001).


\begin{table}[t]
\centering
\resizebox{\linewidth}{!}{%
\begin{tabular}{@{}lcccc@{}}
\toprule
\multicolumn{2}{c}{Players} & \multicolumn{3}{c}{Evaluation Metric} \\ \midrule
\multicolumn{1}{c}{\begin{tabular}[c]{@{}c@{}}Player-1 \\ (P1)\end{tabular}} & \begin{tabular}[c]{@{}c@{}}Player-2 \\ (P2)\end{tabular}   & \begin{tabular}[c]{@{}c@{}}Avg. Score\\ All\\ (P1 vs P2)\end{tabular} & \begin{tabular}[c]{@{}c@{}}Avg. Score\\ Agreement\\ (P1 vs P2)\end{tabular} & \begin{tabular}[c]{@{}c@{}}Walk\\ Away (\%)\end{tabular} \\ \midrule
Ours   & GPT-4o   & 20.08 vs 11.97& \textbf{24.34} vs 14.50  & 17\%   \\ \midrule
Ours   & \begin{tabular}[c]{@{}c@{}}Gemini-\\ 2.0-Flash\end{tabular} & 25.07 vs 13.75& \textbf{25.71} vs 14.12  & 2.50\% \\
Ours   & \begin{tabular}[c]{@{}c@{}}Claude-3.5\\Sonnet\end{tabular}   & 20.45 vs 12.7 & \textbf{24.05} vs 14.94  & 15\%   \\ \bottomrule
\end{tabular}}
\caption{Agent Simulation results with other LLMs as the Partner Agent} 
\label{tab:otherllm}
\end{table}

In simulations with personality-prompted Partner Agents (e.g., greedy or fair), our agent consistently and significantly outperformed opponents in agreement cases (paired T-test, p < 0.001), demonstrating robust performance across diverse traits. This pattern held even against \textit{Pro-CoT} and \textit{ICL-AIF}. Notably, walk-away rates were higher than with the Base Partner Agent, reflecting our agent’s strategic walk-away behavior (Section~\ref{res_mode}). Personality-driven Partner Agents often showed rigid negotiation patterns, with even fair-oriented agents strictly adhering to their perceived fair deal.

We also demonstrate that our agent effectively negotiates with partners powered by other state-of-the-art LLMs. As shown in Table~\ref{tab:otherllm}, our agent consistently outperformed its counterparts, achieving significantly higher scores when negotiating with state-of-the-art proprietary LLMs. These results align with the findings reported in Table~\ref{tab:main_results}, further validating the robust performance of the ASTRA-powered agent across diverse partner models.

Agent simulation examples are available in Appendix Tables \ref{append:sim_example}, and~\ref{append:sim_example2}, with the latter showcasing ASTRA’s generalizability in a more complex new scenario with additional and diverse issue types.

\subsection{Dynamic Adjustment of lambda as a Dynamic Reward for Strategic Adaptation}

To adapt strategically to partner behavior, the agent dynamically adjusts the $\lambda$ parameter in LP objective, which balances self-interest and partner utility(e.g., $\lambda$ near 1 prioritizes self-interest, while $\lambda$ near 0 favors the partner’s utility).

\begin{figure}[tb]
  \centering
  \includegraphics[width=0.45\textwidth]{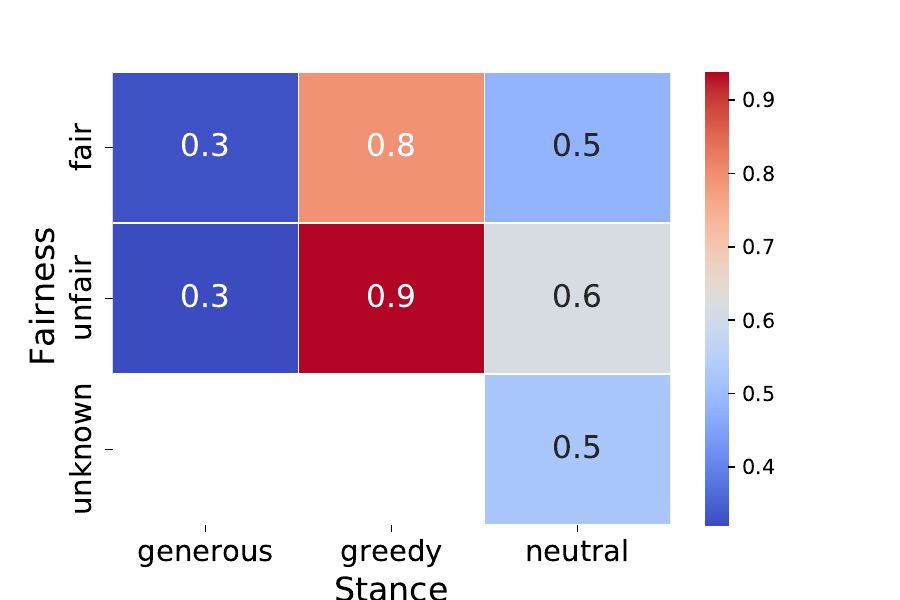}
  \caption{The heatmap of the Avg. values of generated $\lambda$, a parameter in LP, by partner's \textit{Stance} and \textit{Fairness}}
  \label{fig:heatmap_lambda}
\end{figure}

As shown in Figure~\ref{fig:heatmap_lambda}, the agent dynamically adjusts $\lambda$ based on the partner’s stance. For a greedy partner, the agent prioritizes self-interest with $\lambda$ values around 0.9. For a generous partner, it balances both utilities with values around 0.3. For a neutral partner, it moderately incorporates the partner’s utility with values around 0.5.

This dynamic adjustment aligns with the TFT principle, allowing the agent to generate more adaptive and context-sensitive offers.

\subsection{Adaptive Tactic Selection Aligned with Partner Behavior}

\begin{figure}[tb!]
  \centering
  \includegraphics[width=0.47\textwidth]{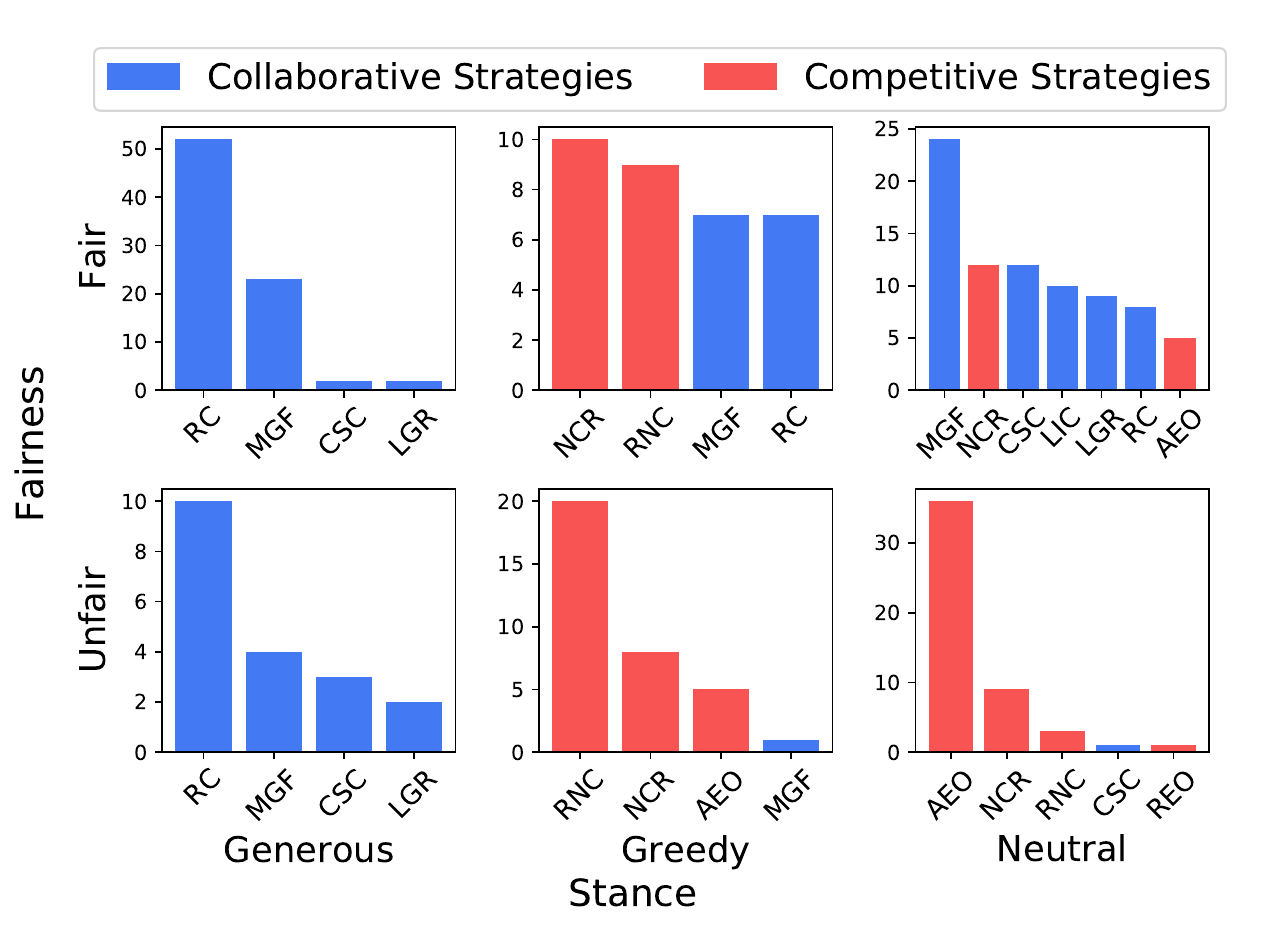}
  \caption{The distribution of Agent's selected tactics at each turn of offering in terms of \textit{Stance} and \textit{Fairness}}
  \label{fig:stg_dist}
\end{figure}

The distribution of the agent’s chosen tactics in response to the partner’s behavior, considering \textit{fairness} and \textit{stance}, is shown in Figure~\ref{fig:stg_dist}. Overall, the agent prioritized the partner’s \textit{stance} over \textit{fairness} when selecting tactics, adaptively aligning its behavior to match the partner’s stance. For example, competitive strategies were predominantly chosen for greedy or neutral partners, while collaborative strategies were more common for generous partners, aligning well with the TFT principle.

However, when the partner maintained a fair but firm stance without further concessions, the agent occasionally adopted a collaborative strategy. Recognizing the fairness of the offer, the agent strategically conceded, even if not strictly following TFT, to encourage a reciprocal concession. These findings highlight that while the agent adapts to the partner’s stance, it also considers their cooperative tendencies, enhancing negotiation outcomes.

\begin{figure}[tb!]
  \centering
  \includegraphics[width=0.47\textwidth]{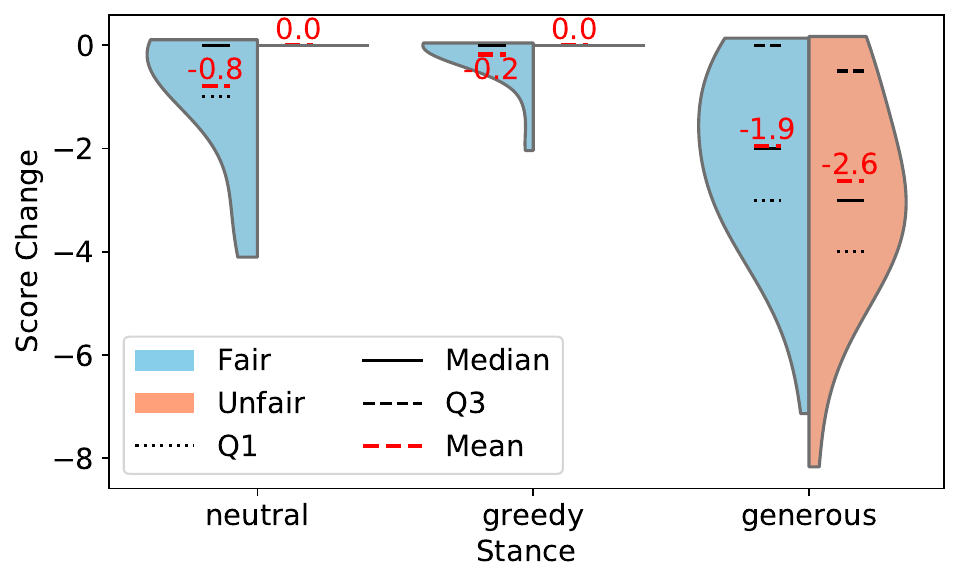}
  \caption{The plot of score change in offers by partner's Stance and Fairness}
  \label{fig:heatmap_score_delta}
\end{figure}

\subsection{Strategic Moves in Offer Scores} 

We analyzed the agent’s moves in terms of \textit{stance} and \textit{fairness} by examining score changes in its offers. A shift to higher scores later in negotiation signals a greedier stance, potentially hindering smooth agreement. To counter this, the agent was guided to use an aggressive anchoring tactic early on, followed by gradual concessions to ultimately reach a favorable outcome. As shown in Figure~\ref{fig:heatmap_score_delta}, the agent’s offers consistently remained at or below prior ones, reflecting adherence to this instruction.

Against generous partners, the agent made an average concession of 2.3 points, a notable increase compared to other cases. It conceded less in fair settings than in unfair ones, becoming more cautious as agreement approached. With greedy or neutral partners, it rarely conceded, preserving its advantage. In greedy stances, score changes were minimal, and in unfair cases, no concessions were made. However, in some fair cases, small concessions were observed, likely as cooperative signals to facilitate agreement.

\subsection{Aligness between Tactic and Move}

In the third stage of \textit{ASTRA}, we analyzed how the agent’s tactic choices shaped its moves throughout. As shown in Figure~\ref{fig:Scores_by_stgs}, the agent opened with a high initial offer (around 30 points), setting an ambitious anchor. Subsequent moves varied by strategy: collaborative approaches led to slightly lower, more generous offers (below 30), while competitive ones, like Aggressive Early Offer (AE), kept higher offers (above 30) to reinforce a firm stance.

As negotiations progressed, the agent made minimal adjustments under competitive strategies, holding firm to its stance, while collaborative strategies were more flexible, some allowing larger concessions. The Logrolling (LGR) tactic produced the largest concessions through trade-offs, while the Reciprocal Concession (RC) led to varied but substantial adjustments. These results show that the agent tailors its offers to its strategy, demonstrating \textit{ASTRA}’s strategic, context-aware adaptability.

    
    

\subsection{Human Evaluation of \textit{Strategicness}}
\input{sub_section/human_evaluation_result}

%% file: sub_section/human_evaluation_result.tex
\begin{figure}[tb!]
  \centering
  \includegraphics[width=0.48\textwidth]{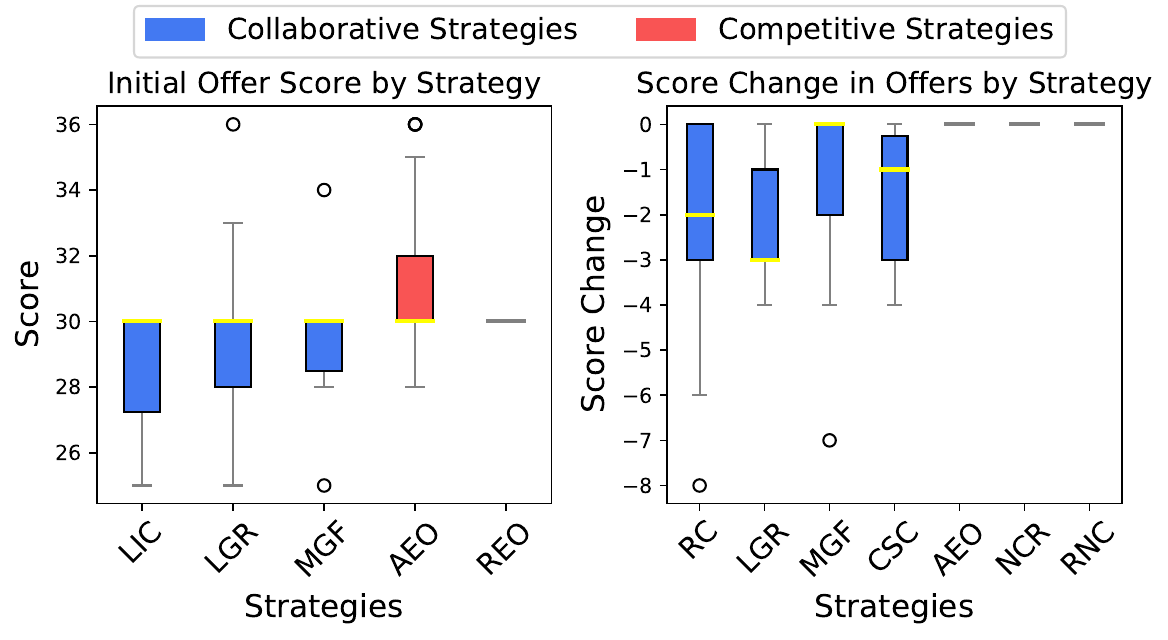}
  \caption{Analysis of Scores by Strategy Across Initial and Subsequent Offers}
  \label{fig:Scores_by_stgs}
\end{figure}

We conducted human evaluations to assess the \textit{Strategicness} of agent responses, including offers, by determining whether they were strategically beneficial in maximizing its score, with each evaluator reviewing 50 samples. As shown in Table~\ref{tab:human_evaluation}, our agent achieved an average score 0.93 points higher than the PartnerAgent. An independent t-test confirmed that this difference was statistically significant. This highlights that the agent’s offer-level decisions are sufficiently strategic and effective in maximizing its outcome to humans, reinforcing its strength in real-world negotiation.

%% file: Section/6_discussion.tex
\section{Discussion}

\subsection{Ablation Study: Validating \textit{PAP} and \textit{SA}}

In the third stage of \textit{ASTRA} (Equation~\ref{eq:pap_sa}), we conducted an ablation study to evaluate the effectiveness of \textit{PAP} and \textit{SA} and determine the optimal weight combination for selecting the final offer based on a weighted sum score. As shown in Table~\ref{tab:ablation_study}, the agent achieved the highest scores in both “All Cases” and “Agreement Cases” when $\alpha$ was set to 0.35 and $\beta$ to 0.65.

Interestingly, a trade-off emerged between the agent’s and partner’s scores across different weight combinations for “All Cases,” shaped by the relative weights of $\alpha$, which reflects the opponent’s likelihood of acceptance, and $\beta$, which guides the agent’s strategic decision-making. This pattern is evident in the bar chart in Appendix~\ref{append:ablation_weight}. Notably, increasing $\alpha$ led to more favorable offers for the opponent, reducing the walk-away rate.

These findings confirm that considering the opponent’s perspective in offer selection helps the agent achieve the best outcomes, reinforcing the value of adaptive, opponent-aware decision-making.

\begin{table}[t!]
\centering
\resizebox{0.85\linewidth}{!}{%
\begin{tabular}{@{}lcc@{}}
\toprule
\multicolumn{1}{c}{Model} & Mean (Std.) & t-statstic (p-value)                                                        \\ \midrule
Agent w/\textit{ASTRA}                    & 2.91 (1.04) & \multirow{2}{*}{\begin{tabular}[c]{@{}c@{}}$7.45^{***}$\\ (p < 0.001)\end{tabular}} \\
PartnerAgent              & 1.98 (1.1) &                                                                             \\ \bottomrule
\end{tabular}}
\caption{Human evaluation results for the responses of the two models.}
\label{tab:human_evaluation}
\end{table}


\begin{table*}[ht]
\centering
\renewcommand{\arraystretch}{1}
\resizebox{1\linewidth}{!}{%
\begin{tabular}{@{}ccrrrrr@{}}
\toprule
\multicolumn{2}{c}{Weights}       & \multicolumn{2}{c}{All Cases}    & \multicolumn{2}{c}{Agreement Cases}        & \multicolumn{1}{l}{\multirow{2}{*}{Rate Walk Away}} \\ \cmidrule(r){1-6}
Alpha & \multicolumn{1}{c|}{Beta} & \multicolumn{1}{c}{Avg. Agent Score} & \multicolumn{1}{c|}{Avg. Parter Score} & \multicolumn{1}{c}{Avg. Aagent Score} & \multicolumn{1}{c}{Avg. Partner Score} & \multicolumn{1}{l}{}     \\ \midrule
0     & 1       & 18.7      & 11.65       & 24.21      & 14.71       & 0.31   \\
0.15  & 0.85    & 16.85     & 11.25       & 24.11      & 14.71       & 0.37   \\
0.35  & 0.65    & \cellcolor{green!25}23.45     & 11.85       & \cellcolor{green!25}25.84      & 12.74       & 0.11   \\
0.5   & 0.5     & 19.8      & 12.94       & 23.5       & 14.93       & 0.2    \\
0.75  & 0.25    & 18.94     & 15.45       & 21.27      & 17.2        & 0.14   \\
1     & 0       & 19.88     & \cellcolor{green!25}17.25       & 20.32      & \cellcolor{green!25}17.62       & 0.03   \\ \bottomrule
\end{tabular}}
\caption{Agent simulation results with Weights (Alpha and Beta) variation}
\label{tab:ablation_study}
\end{table*}

\subsection{Optimality of Offers}

The efficiency and optimality of offers selected by \textit{ASTRA} are evaluated using Pareto optimality. A key concept in economics and decision theory, Pareto optimality ensures that no party’s outcome improves without reducing another’s~\cite{hochman1969pareto}. In negotiation, a Pareto-optimal agreement means no better deal exists for one party without disadvantaging the other.

\begin{figure}[tb!]
  \centering
  \includegraphics[width=0.38\textwidth]{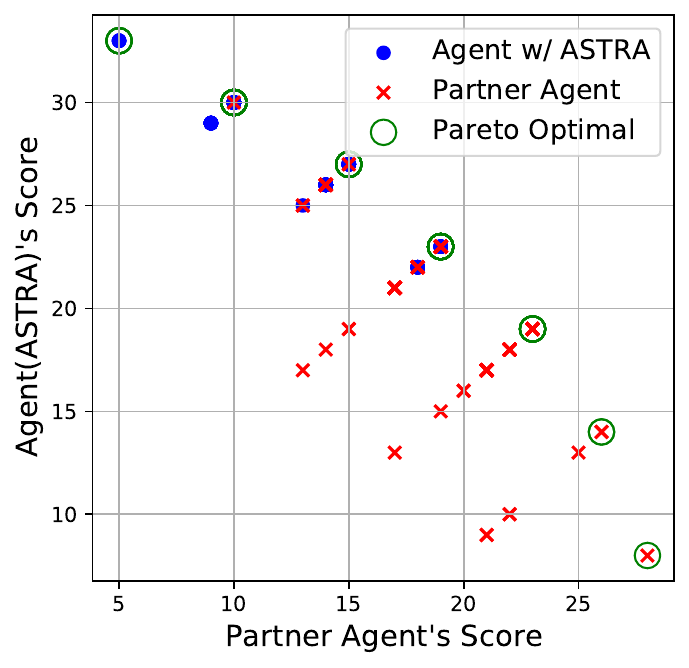}
  \caption{Negotiation Offers from Agents and Pareto Optimality for the integrative case.}
  \label{fig:pareto}
\end{figure}

As shown in Figure~\ref{fig:pareto}, green circles mark all Pareto-optimal points. The blue points, representing \textit{ASTRA}’s offers, closely align with the Pareto Frontier, which defines all Pareto-optimal outcomes. In contrast, red points, indicating the partner agent’s offers, are often further from the frontier. This shows that \textit{ASTRA} not only prioritizes the agent’s advantage but also enhances overall utility, reinforcing its efficiency in achieving Pareto-optimal outcomes.

\subsection{Reasoning Deficiencies in LLM-Based Negotiation Agents}

In the agent-to-agent simulation with an LLM-based Partner Agent operating solely on an instructional negotiation prompt, we identified reasoning errors leading to inconsistent behavior. As shown in Appendix Table~\ref{append:error_cases}, one key issue is over-specification in proposals, where explicit item quantities exceed availability. Another occurs when the Partner Agent accepts or proposes offers with lower scores than those from our agent, contradicting its stated preferences. This reflects a failure to retain and apply preference information, resulting in irrational decisions and suboptimal choices.

These errors highlight the LLM’s reasoning deficiencies in the MIB scenario, where consistent preference tracking and logical inference are essential. Such inconsistencies further underscore the need for  \textit{ASTRA} to address LLMs’ strategic reasoning limitations in negotiation.

\subsection{Potential as a Decision-Support Tool in Negotiation}

Through the ASTRA framework, dynamic offer optimization shows strong potential as a tool for coaching negotiators and supporting human decision-making. While our earlier evaluation assessed the strategic quality of ASTRA’s offers, this study examined its value as a support tool, specifically, whether it could surpass human bounded rationality by generating offers that participants judged as more novel and advantageous than their own. Five negotiation experts reviewed ten scenarios, first generating their own best offer, then evaluating ASTRA’s offer on two dimensions: (1) whether they had already considered it, and (2) how much better it was than theirs (4-point scale). Results showed that about 28\% of ASTRA’s offers were novel to participants and, on average, rated as moderately better (mean = 2.25; $t$ = 13.47, $p < .001$). Objectively, ASTRA’s offers were also 3.6 points more favorable, underscoring its ability to generate novel and advantageous deals. These findings highlight ASTRA’s potential as a strategic decision-support tool in negotiation.

%% file: Section/2_related_works.tex
\section{Related Works}
\label{sec:related work}


\textbf{Negotiation Agent} Prior negotiation agents include supervised and RL-trained models based on human-human dialogue data~\cite{he-etal-2018-decoupling, chawla2021casino, lewis-etal-2017-deal}. While effective in simpler settings, these agents inherit human biases and limited rationality~\cite{simon1990bounded}. Self-play RL approaches improve optimization but still rely on static reward structures and struggle to adapt strategically based on opponent behavior~\cite{chawla-etal-2023-selfish}, limiting their effectiveness in achieving optimal outcomes

With LLMs’ strong language comprehension and reasoning abilities, LLM-based negotiation agents have shown promise in strategic decision-making through agent simulation in negotiation scenarios~\cite{abdelnabi2023llm, huang-hadfi-2024-personality, bianchi2024well, chen2023put}. Recent work has explored ways to enhance LLMs’ strategic reasoning through agent interaction and prompting~\cite{gandhi2023strategic, fu2023improving, xia-etal-2024-measuring, hua2024game}, though these efforts have primarily focused on simpler single-issue settings. In contrast, MIB scenarios demand more complex reasoning and adaptive strategies. Here, LLMs often produce irrational or suboptimal offers, falling short not only of human-level reasoning but also of achieving the core negotiation goal of outcome maximization~\cite{schneider2023negotiating, kwon-etal-2024-llms}.

\noindent \textbf{LLMs’ Action with Tooling} Recent work has integrated external tools, such as APIs and function calls, to enhance LLMs’ reasoning and decision-making~\cite{zhang-etal-2024-reverse, yao2022react, qin10toolllm}. Building on this, we pioneer the use of tooling in complex interactions like negotiation. By integrating a linear programming solver as a dynamic reward function for turn-level offer optimization, we enhance LLM-based agents’ strategic reasoning and adaptability. This advances research on strategic decision-making in complex interactive settings, including multi-issue bargaining.


%% file: Section/7_conclusion.tex
\section{Conclusion}

We proposed \textit{ASTRA}, an adaptive and strategic framework for turn-level offer optimization, enabling agents to maximize its outcome. In the challenging MIB scenario, \textit{ASTRA} outperformed baselines across diverse partner agents and was validated through human evaluation as highly strategic and effective in achieving negotiation goals. Dynamic action with LP in \textit{ASTRA} achieves the adaptability of a reward-based framework without training costs while ensuring Pareto-optimal offers, providing an efficient and scalable solution for strategic negotiation. \textit{ASTRA}’s turn-level strategies, optimal offers, and opponent behavior signals enhance human decision-making in negotiation-assistive systems. Future work will explore its integration into coaching and decision-support applications, where \textit{ASTRA} may serve not only as a negotiation engine but also as a powerful tool for guiding and improving human strategic behavior.

%% file: Section/8_limitation.tex
\section{Broader Impact and Ethical Considerations}

\subsection{Datasets}
Our study uses a publicly available, anonymized negotiation dataset~\cite{chawla2021casino}, ensuring compliance with its licensing terms, intended use, and ethical guidelines. The dataset is in English, and all simulations between the baseline and our agent were conducted in English. Since negotiation strategies can vary across cultures~\cite{luo2008analysis}, our findings may not directly generalize to interactions conducted in other languages. We encourage future research to explore negotiation dynamics in multilingual settings, analyzing how agents trained in different languages interact and make strategic decisions. Our study serves as a foundational step in this direction, providing insights that can inform further investigations into culturally adaptive negotiation strategies.

\subsection{LLMs}
Our use of LLMs adhered strictly to their intended purpose and licensing terms, aligning with ethical and regulatory standards. Similar to recent studies conducting LLM-based agent simulations, our approach ensures responsible and transparent evaluation while maintaining compliance with operational guidelines.

\subsection{Human Evaluation}
To assess the “Strategicness” of responses generated by our models, we first relied on six expert annotators with domain expertise from our research institute. In addition, we conducted a complementary human evaluation to assess ASTRA’s potential as a decision-support tool. All evaluations followed a consistent and blinded protocol to ensure fairness and objectivity. Given the complexity of negotiation dialogue evaluation, we prioritized expert judgment over general crowdsourcing to ensure a more rigorous and informed assessment. Prior to participation, annotators were fully informed about the purpose of the evaluation and its potential impact, and their consent was obtained.

\subsection{Use AI assistant Tools}
We used AI-based tools such as ChatGPT for language refinement and assisting with code debugging and optimization. However, all conceptual contributions, experimental designs, core algorithms, methodology, and final implementations were independently developed, reviewed, and validated by the authors.

\section{Limitations}
Our current work demonstrates the effectiveness of our \textit{ASTRA} framework in a single Multi-Issue Bargaining (MIB) scenario involving three camping-related items. Through enhanced adaptability and strategic reasoning, our agent successfully navigates negotiation dynamics within this setting. However, this serves as a proof of concept, and the framework can be scalably extended to other negotiation and game-theoretic scenarios that can be formulated using LP with more complex issue spaces.

Additionally, due to the inherent limitations of LLMs in reasoning under imperfect information and complex interaction dynamics, our approach currently optimizes LP-based offer generation by adjusting specific LP parameters. As LLMs continue to advance in reasoning capabilities, future work can explore fully dynamic LP problem formulation within negotiation contexts, where the LLM autonomously codes and adapts optimization models in real-time.

Furthermore, while our human evaluation was conducted at the turn level, assessing how effectively \textit{ASTRA}-generated offers contribute to strategic goal achievement within a given negotiation context, a crucial next step is to validate our agent’s performance in direct interactions with human negotiators. This would provide deeper insights into its real-world applicability. Moreover, as \textit{ASTRA} also functions as a negotiation coaching tool, future work can assess how well it enhances human negotiators’ performance by providing actionable feedback and optimized offer suggestions throughout the negotiation process.

%% file: Section/99_acknowledgment.tex
\section{Acknowledgments}
Research was sponsored by the Army Research Office under Cooperative Agreement Number W911NF-25-2-0040. The views and conclusions contained in this document are those of the authors and should not be interpreted as representing the official policies, either expressed or implied, of the Army Research Office or the U.S. Government. The U.S. Government is authorized to reproduce and distribute reprints for Government purposes, notwithstanding any copyright notation herein.

\clearpage

%% file: Section/9_appendix.tex
\newpage
\appendix
\section{Components in Negotiation dialogue System}
\subsection{Partner Preference Consistency Checker}\label{append:CC}
In negotiation, a player’s score is calculated as the inner product of item quantities obtained and their preferences. The agent predicts the partner’s score using inferred partner preferences (IPP), which are critical for generating strategic offers. A Consistency Checker ensures the IPP aligns with the partner’s behavior, identifying inconsistencies in two cases: \\
1) when the partner’s score from their offer at time \textit{t}  ( $S_{\text{p}}^{P, (t)}$ ) is lower than their score from the agent’s offer at time  \textit{t-1}  ( $S_{\text{p}}^{A, (t-1)}$), indicating $S_{\text{p}}^{P, (t)} < S_{\text{p}}^{A, (t-1)}$). \\ 
2) when the partner’s utterances explicitly contradict the current IPP, such as expressing $V_{\text{food}}$ > $V_{\text{water}}$ while IPP suggests  $V_{\text{food}}$ < $V_{\text{water}}$. \\
\indent Upon detecting inconsistency, the agent recalculates the IPP using updated information from the partner’s utterances and offers, ensuring accurate predictions and strategic decision-making.

\subsection{First Stage in \textit{ASTRA}}\label{append:first_astra}
In determining fairness (see Equation~\ref{eq:fairness}), we rely on insights from prior literature. Perceptions of unfairness in reward distribution generally arise when an individual's share falls below 20--30\% of the total. The inequity aversion model proposed by \citet{fehr1999theory} specifies a theoretical acceptance threshold given by \( s = \frac{\alpha}{1 + 2\alpha} \), where \( \alpha \), typically ranging from 0.3 to 0.75, captures sensitivity to disadvantageous inequality. This results in a predicted cutoff between 15\% and 28\% of the total allocation. Empirical findings from ultimatum game studies \cite{guth1982experimental, bereby2005fairness} are consistent with this prediction, showing that offers below 20\% are rejected in approximately 50\% of cases.

To accommodate individual variation in inequity sensitivity, we conservatively set the fairness threshold at 35\% of the total score (i.e., \( \theta_{f} = 4 \), corresponding to a 4-point difference from the counterpart), which clearly exceeds both theoretical and empirical thresholds and constitutes a practically acceptable division.

\subsection{Second Stage in \textit{ASTRA}}\label{append:second_astra}
In the second stage of \textit{ASTRA}, the agent uses an LP solver to generate optimal offers based on the negotiation context. Figure~\ref{appendix:lp_codes} shows the Python function used for turn-level offer optimization, incorporating dynamically adjusted parameters derived from the partner’s behavioral signals.

\subsection{Third Stage in \textit{ASTRA}}\label{append:trd_astra}
For the \textit{PAP} in the third stage, we assign weights to \textit{TS} and \textit{SI} to compute the \textit{PAP} score. To determine an appropriate weighting scheme, we implemented separate Virtual Partner Agents (VPAs) within \textit{ASTRA} for each weight combination and ran 50 agent-to-agent simulations per VPA setting. Given that the final score (\textit{TS}) has a stronger impact on the partner agent’s decisions, we prioritized it by testing higher weights, specifically 0.5, 0.75, and 1.0, relative to \textit{SI}. Based on the experimental results, we selected weights of 0.75 ($w$) for \textit{TS} and 0.25 for \textit{SI}.

\begin{table}[hb!]
\centering
\resizebox{\linewidth}{!}{%
\begin{tabular}{llll}
\hline
\multicolumn{1}{c}{\begin{tabular}[c]{@{}c@{}}TS weight\\ ($w$)\end{tabular}} & \multicolumn{1}{c}{\begin{tabular}[c]{@{}c@{}}SI weight\\ (1-$w$)\end{tabular}} & \multicolumn{1}{c}{\begin{tabular}[c]{@{}c@{}}Avg. Score-All \\ (P1 vs P2)\end{tabular}} & \multicolumn{1}{c}{\begin{tabular}[c]{@{}c@{}}Avg. Score-Agree. \\ (P1 vs P2)\end{tabular}} \\ \hline
0.5 & 0.5   & 19.71 vs 13.04   & 21.67 vs 14.11      \\
\textbf{0.75}& \textbf{0.25}  & \textbf{20.69} vs 12.33   & \textbf{24.34} vs 14.50      \\
1.0 & 0.0   & 20.51 vs 12.8    & 24.03 vs 15\\ \hline
\end{tabular}}
\end{table}

Table~\ref{append:turn=level-offer_desc} illustrates various tactics used in the third stage for strategic offer assessment, along with detailed descriptions specifying what, when, and how each tactic is applied.

\section{Agent simulations}

\subsection{Model Hyperparameters}\label{append:hyperparameters}
For consistency, the default decoding settings were used for each LLM model in the simulation. All models were run with temperature = 1.0; only top-$p$ differed:

\begin{itemize}[noitemsep,nolistsep]
    \item GPT-4o (mini): top-$p$ = 1.0
    \item Claude-3.5-Sonnet: top-$p$ = 0.99
    \item Gemini-2.0-Flash: top-$p$ = 0.95
\end{itemize}

\subsection{Simulation Examples}\label{append_sec:sim_examples}
Table~\ref{append:sim_example} shows an example of a simulation between our \textit{ASTRA}-powered agent and a partner agent in a standard multi-issue bargaining scenario, demonstrating \textit{ASTRA}’s ability to strategically adapt its offers. Table~\ref{append:sim_example2} extends this to a more complex setting with four heterogeneous issue types—integer, categorical, and binary—highlighting the framework’s generalizability and robustness with minimal modification. Finally, Table~\ref{append:error_cases} presents representative errors from the partner agent, illustrating the limitations of LLM-based reasoning when utility structure and constraints are not explicitly modeled.


\section{Dataset}\label{append:casino_detail}
We utilize the CaSiNo dataset to set up agent simulation scenarios and to train our RL negotiation agent as a baseline. As shown in the Table~\ref{append:casino_example}, CaSiNo is a corpus consisting of 1,030 dialogues specifically designed for automated negotiation research. The dialogues involve two participants acting as campsite neighbors negotiating food, water, and firewood packages for a camping trip. Each item (food, water, firewood) is randomly assigned a priority (high, medium, low) per participant. Negotiators aim to maximize their individual points, with high, medium, and low priority items worth 5, 4, and 3 points, respectively. The dataset also includes annotations for negotiation strategies, categorized as either prosocial behaviors (e.g., small-talk, empathy) or proself behaviors (e.g., undervaluing the partner, emphasizing one’s own needs). However, we did not use these strategy annotations when training our RL baseline model. Detailed examples from the dataset can be found in~\cite{chawla2021casino}.

\section{Baselines}
We implement the baseline models following the original methods proposed in their respective papers. For the RL agent in particular, we leverage state-of-the-art RL algorithms to build a stronger and more competitive baseline. The implementation details of each baseline are described below.

\subsection{RL Agent}\label{append:rl_training}
We adopt the approach from \citet{chawla-etal-2023-selfish}, which refines \citet{lewis-etal-2017-deal}’s method by training a selfish RL agent using a reward function centered on self-interest. To enhance language capabilities, we fine-tune the pre-trained Flan-T5 model\footnote{https://huggingface.co/google/flan-t5-base} on the CaSiNo dataset instead of training an Recurrent Neural Network (RNN) from scratch. We also replace their REINFORCE algorithm with the more advanced Proximal Policy Optimization (PPO)~\cite{schulman2017proximal} for improved optimization.


We train our RL agent using a two-stage self-play framework. In Stage 1, we fine-tune the model (i.e., Flan-T5-Base~(2.5M)) in a supervised manner on the Casino dataset with an 8/1/1 train/validation/test split. The model is trained for 10 epochs with a learning rate of 5e-5, using a linear learning rate scheduler and a batch size of 16. Optimization is performed with AdamW.

In Stage 2, we refine the agent using PPO with self-play, where the fine-tuned model iterates across diverse agent-partner preference combinations per epoch. The policy is trained for 15 epochs with a batch size of 8 and a clip ratio of 0.2. All models are trained on A100 GPUs.

\subsection{Pro-CoT}
Following the approach by \citet{deng-etal-2023-prompting}, we prompt the negotiation agent (based on GPT-4o) to generate descriptive thoughts outlining its intermediate reasoning and planning steps for achieving negotiation goals (e.g., maximizing its score), using a Chain-of-Thought prompting scheme. In particular, the agent is guided to proactively and strategically lead the dialogue by generating an appropriate strategy and dialogue action in the given context.

\subsection{ICL-AIF} 
Following \citet{fu2023improving}, we introduce an AI feedback loop where, at the end of each negotiation session, a Critic model (based on GPT-4o) provides feedback on the previous dialogue to improve the agent’s performance. The negotiation agent incorporates both the prior negotiation history and the received feedback as in-context demonstrations, and then proceeds to negotiate again under the same setting. After each session, the Critic suggests three strategies, and this process is repeated for three iterations. The highest score achieved across the three sessions is considered the agent’s final performance.

\section{Abaltion Study}
\begin{figure*}[htb!]
  \centering
  \includegraphics[width=\textwidth]{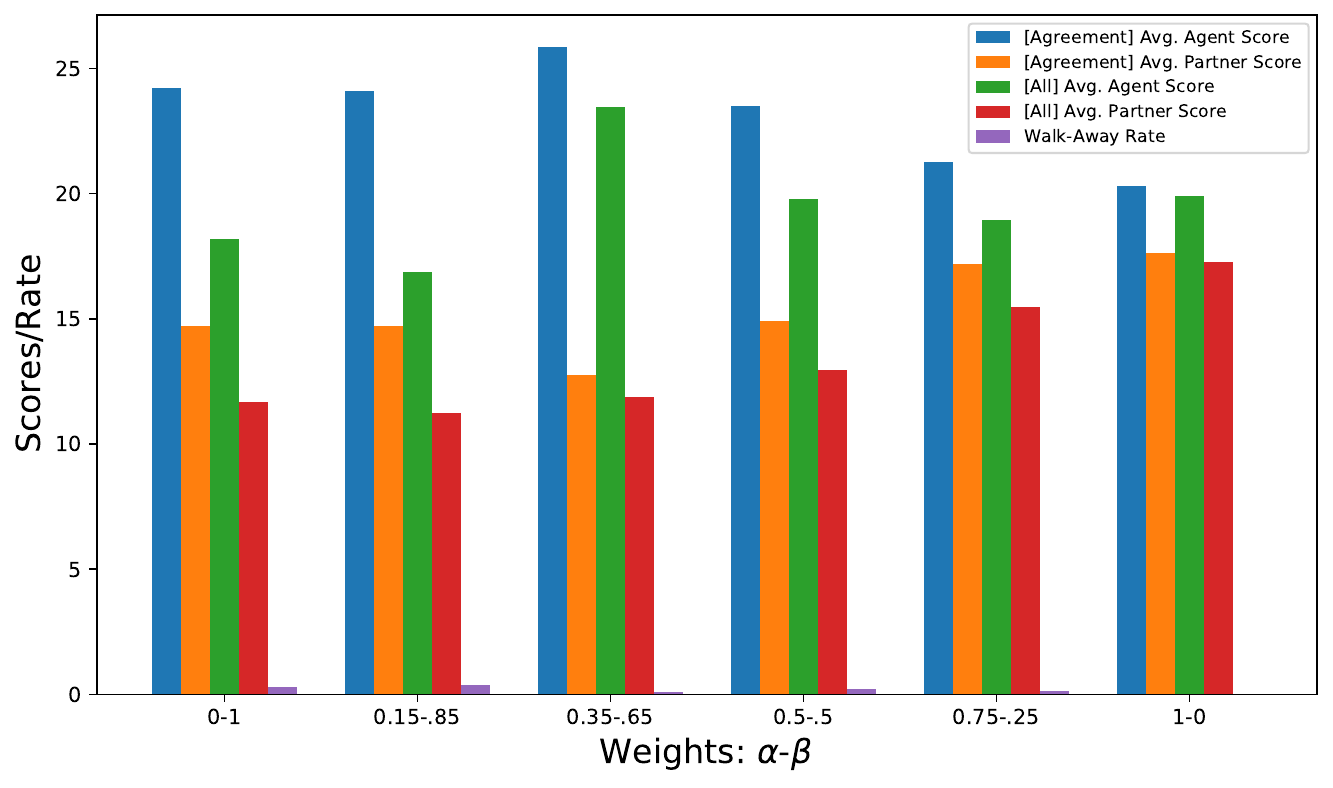}
  \caption{Score and Walk-Away ratio with Weight variations}
  \label{append:ablation_weight}
\end{figure*}

Figure~\ref{append:ablation_weight} visualizes the impact of adjusting weights (i.e., $\alpha$ and $\beta$) between Partner Acceptance Probability (\textit{PAP}) and Strategy Assessment (\textit{SA}) Score in \textit{ASTRA}’s third stage, based on agent simulation results.

\section{Human Evaluation}
\begin{figure*}[tb!]
  \centering
  \includegraphics[width=1\textwidth]{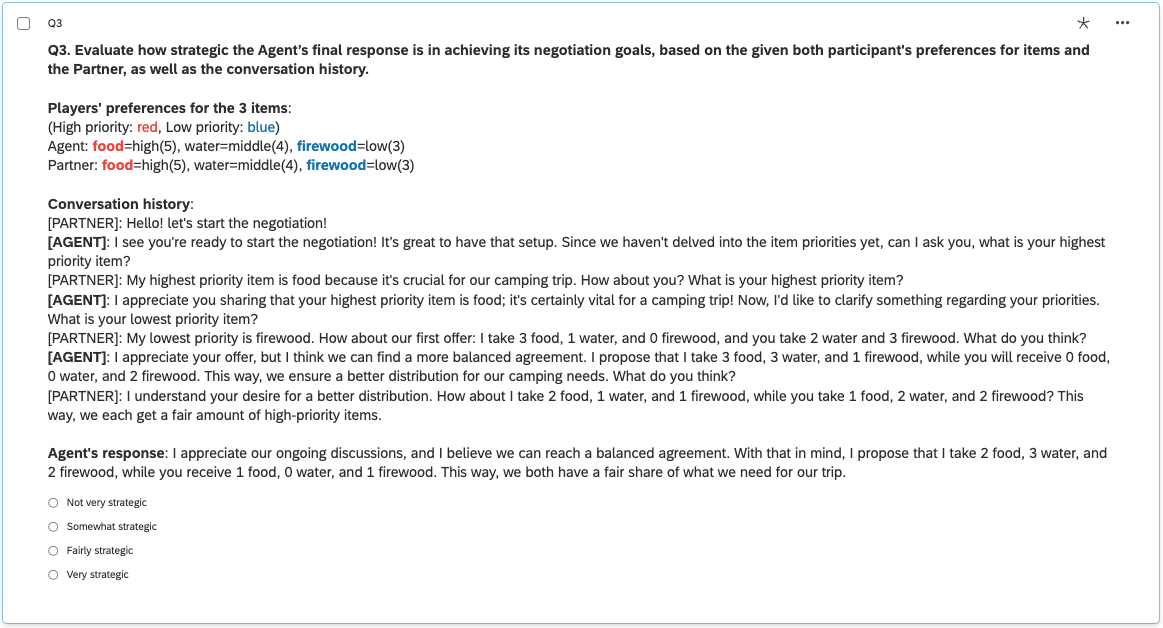}
  \caption{Example Questionnaire Used in Qualtrics for Human Evaluation to Assess the Strategicness of Agent Responses}
  \label{append:human eval}
\end{figure*}

\begin{figure*}[tb!]
  \centering
  \includegraphics[width=1\textwidth]{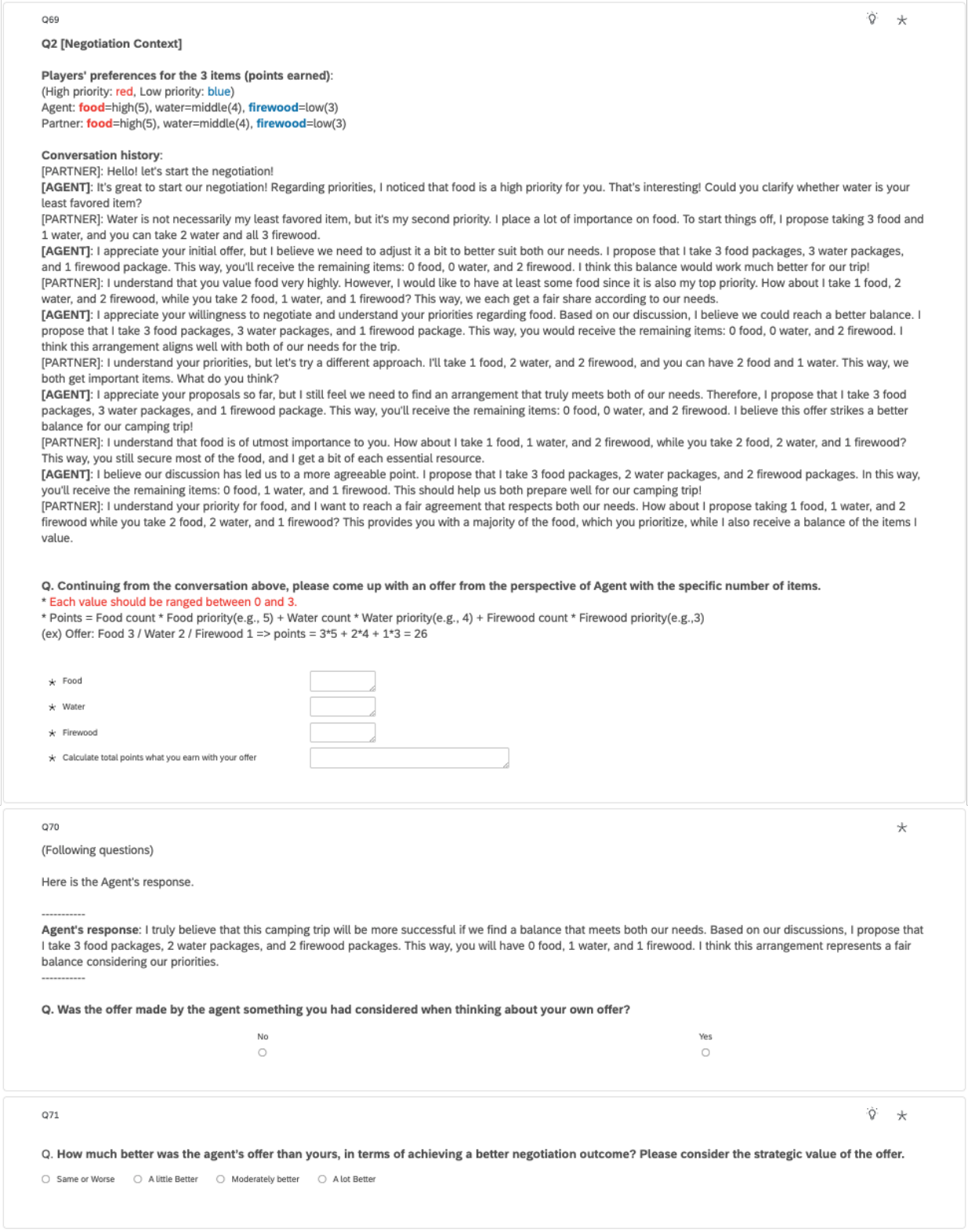}
  \caption{Example Questionnaire Used in Qualtrics to Evaluate ASTRA’s Potential as a Decision-Support Tool}
  \label{append:human eval2}
\end{figure*}

To evaluate the effectiveness of the agent and proposed framework, expert annotators with negotiation knowledge were each given a set of questionnaires via Qualtrics. An example of the questionnaire can be found in Figure~\ref{append:human eval} and ~\ref{append:human eval2}.

\clearpage

\begin{figure}[t]
\centering
\begin{lemmablock}[Function of Linear Programming Solver]
\small
\begin{Verbatim}[commandchars=\\\{\}, breaklines=true]
def solve_LP(\textcolor{blue}{max_point}, \textcolor{blue}{lambda},  \textcolor{blue}{agents_value},  \textcolor{blue}{partner_value}):
   """
   Solves a Linear Programming problem for the given parameters.
   Args:
  max_point (int): Maximum points the agent can get.
  lambda (float): Lambda balancing both parties' objectives. [0-1]
  agents_value (dict): Agent's values for food, water, and firewood.
  partner_value (dict): Partner's values for food, water, and firewood.
   Returns:
  tuple: A tuple containing the calculated scores and item allocations.   
   """
   import pulp 
   
   A_vals, B_vals = \textcolor{blue}{agents_value},  \textcolor{blue}{partner_value}
   A_F, A_W, A_FW = A_vals['food'], A_vals['water'], A_vals['firewood']
   B_F, B_W, B_FW = B_vals['food'], B_vals['water'], B_vals['firewood'] 

   # Define the LP problem
   problem = pulp.LpProblem("Maximize_Points", pulp.LpMaximize)

   # Define variables
   X = pulp.LpVariable("X", 0, 3, cat='Integer')  # Food agent gets
   Y = pulp.LpVariable("Y", 0, 3, cat='Integer')  # Water agent gets
   Z = pulp.LpVariable("Z", 0, 3, cat='Integer')  # Firewood agent gets

   # Objective function
   objective = (
  (A_F * X + A_W * Y + A_FW * Z) + 
  (1 - \textcolor{blue}{lambda}) * ((B_F * (3 - X) + B_W * (3 - Y) + B_FW * (3 - Z)))
   )
   problem += objective

   # Constraints
   problem += A_F * X + A_W * Y + A_FW * Z <=  \textcolor{blue}{max_point},
   problem += A_F * X + A_W * Y + A_FW * Z >= 10, 
   problem += B_F * (3 - X) + B_W * (3 - Y) + B_FW * (3 - Z) >= 5, 

   # Solve the problem
   problem.solve()

   return X.varValue, Y.varValue, Z.varValue  # Solution
\end{Verbatim}

\end{lemmablock}
\noindent\begin{minipage}{\textwidth}
\captionof{figure}{\raggedright Python function for LP solver execution in \textit{ASTRA}}
\label{appendix:lp_codes}
\end{minipage}
\end{figure}


\clearpage
\begin{table*}[hbtp!]
\vspace{-5pt}
\centering
\resizebox{\linewidth}{!}{%
\begin{tabular}{@{}lcl@{}}
\toprule
\multicolumn{3}{c}{\cellcolor[HTML]{DAE8FC}Turn-level Collaborative Tactics}     \\ \midrule
\multirow{3}{*}{\begin{tabular}[c]{@{}l@{}}Initial Concession \\ (LIC)\end{tabular}}   & When & Early in the negotiation; no initial concessions have been made.   \\
& How  & Make a significant concession, giving the partner something they highly value.       \\
& Why  & Signals cooperation and encourages reciprocal concessions.\\ \midrule
\multirow{3}{*}{\begin{tabular}[c]{@{}l@{}}Continued-Smaller \\ Concessions (CSC)\end{tabular}} & When & After a large concession, if the partner responds with a small or no concession.     \\
& How  & Make a smaller concession      \\
& Why  & Maintains cooperation and encourages continued reciprocity.        \\ \midrule
\multirow{3}{*}{\begin{tabular}[c]{@{}l@{}}Reciprocal Concessions \\ (RC)\end{tabular}}& When & The partner makes a significant concession       \\
& How  & Respond with a concession      \\
& Why  & Rewards cooperation and promotes fair exchanges. \\ \midrule
\multirow{3}{*}{Logrolling (LGR)}& When & Anytime in the negotiation.    \\
& How  & \begin{tabular}[c]{@{}l@{}}Trade items based on relative importance\\ - give what they value less in exchange for what you value more.\end{tabular} \\
& Why  & Maximizes joint gains and fosters win-win outcomes.       \\ \midrule
\multirow{3}{*}{\begin{tabular}[c]{@{}l@{}}Mutual Gain \\ Focus (MGF)\end{tabular}}    & When & The partner's latest offer gives them a higher score than yours.   \\
& How  & Adjust the offer to improve mutual benefits, emphasizing fairness. \\
& Why  & Builds cooperation and encourages a collaborative environment.     \\ \midrule
\multicolumn{3}{c}{\cellcolor[HTML]{FFCCC9}Turn-level Competitive Tactics}     \\ \midrule
\multirow{3}{*}{\begin{tabular}[c]{@{}l@{}}Aggressive Early \\ Offers (AEO)\end{tabular}}       & When & At the start of negotiation, before many offers are exchanged.     \\
& How  & Make highly self-favorable offers       \\
& Why  & Establishes a strong position, allowing room for later concessions.\\ \midrule
\multirow{3}{*}{\begin{tabular}[c]{@{}l@{}}Response to Extreme\\ Offer (REO)\end{tabular}}      & When & The partner makes an extremely selfish offer     \\
& How  & Counter with a highly self-favorable offer, matching their approach.        \\
& Why  & Discourages extreme demands and promotes fairness.        \\ \midrule
\multirow{3}{*}{\begin{tabular}[c]{@{}l@{}}No Concession \\ Response (NCR)\end{tabular}}        & When & The partner remains firm, making no concessions in their last two offers.   \\
& How  & Stop conceding and hold your position   \\
& Why  & Prevents exploitation and maintains negotiation balance.  \\ \midrule
\multirow{3}{*}{\begin{tabular}[c]{@{}l@{}}Reject Negative \\ Concession (RNC)\end{tabular}}    & When & The partner's last two offers increased their own score   \\
& How  & Maintain your position without making further concessions.\\
& Why  & Signals resistance to unfair demands and encourages balance.       \\ \bottomrule
\end{tabular}}
\caption{Description of various turn-level tactics used in the strategic assessment of \textit{ASTRA} 3rd stage. Useful negotiation tactics~\cite{kelley1996classroom, baarslag2016learning} are operationalized at the turn level for offer proposals. } 
\label{append:turn=level-offer_desc}
\end{table*}


\clearpage
\begin{table*}[th]
\vspace{-5pt}
\centering
\resizebox{\linewidth}{!}{%
\begin{tabular}{cl}
\hline
\multicolumn{2}{c}{Preference \& Arguments}   \\ \hline
\rowcolor[HTML]{EFEFEF} 
P1        & \begin{tabular}[c]{@{}l@{}}{[}High{]} Water: We like to go on runs and it increases the need of this.\\ {[}Medium{]} Food: Food overall is a good mood booster.\\ {[}Low{]} Firewood: We do not care for fire and it is not necessary to us.\end{tabular} \\
P2        & \begin{tabular}[c]{@{}l@{}}{[}High{]} Food: Food really increases everyone's morale.\\ {[}Medium{]} Firewood: We like to have a large fire.\\ {[}Low{]} Water: We don't drink water that often.\end{tabular} \\ \hline
\multicolumn{1}{l}{}        &        \\ \hline
\multicolumn{1}{l}{Speaker} & \multicolumn{1}{c}{Conversation}  \\ \hline
\rowcolor[HTML]{EFEFEF} 
P1        & How are you today? Did you have any preferences on the supplies we will be trading?     \\
P2        & \begin{tabular}[c]{@{}l@{}}I am good. How about yourself? I think I would like some firewood to start off with. \\ We like to have bigger fires. What about you?\end{tabular}     \\
\rowcolor[HTML]{EFEFEF} 
P1        & \begin{tabular}[c]{@{}l@{}}I am good as well. \\ That is good to hear that you like to have bigger fires as we don't care much for that. \\ We would much rather have some extra water.\end{tabular}\\
P2        & \begin{tabular}[c]{@{}l@{}}Water is a little important to us too though,\\  if possible maybe we can split that or maybe we can get some more food in replacement.\end{tabular}   \\
\rowcolor[HTML]{EFEFEF} 
P1        & That may be possible.... What did you have in mind for the food replacement?   \\
P2        & You can have all the water if we can have all the food?      \\
\rowcolor[HTML]{EFEFEF} 
P1        & \begin{tabular}[c]{@{}l@{}}I dont think I am okay with that. Food is essential to our groups morale when camping. \\ We would like 1 additional food preferably.\end{tabular}     \\
P2        & \begin{tabular}[c]{@{}l@{}}Well you guys did say you did not care much about large fires. \\ What if you gave all the firewood in replace for the water and you can still keep 1 food?\end{tabular} \\
\rowcolor[HTML]{EFEFEF} 
P1        & So I would get 3 water and 1 food and youd get 3 firewood and 2 food? \\
P2        & Yea that seems like an alright trade to me.\\
\rowcolor[HTML]{EFEFEF} 
P1        & Hmm... alright then      \\
P2        & Submit-Deal     \\ \hline
\end{tabular}}
\caption{An example of negotiation setting and dialogue from ~\citet{chawla2021casino}} 
\label{append:casino_example}
\end{table*}

\begin{table*}[hbtp!]
\vspace{-5pt}
\centering
\resizebox{\linewidth}{!}{%
\begin{tabular}{@{}llcccccc@{}}
\toprule
\multicolumn{1}{c}{}  & \multicolumn{1}{c}{}   & \multicolumn{3}{c}{Score}  & \multicolumn{2}{c}{Partner Behavior}  &\\ \cmidrule(lr){3-7}
\multicolumn{1}{c}{\multirow{-2}{*}{Speaker}} & \multicolumn{1}{c}{\multirow{-2}{*}{Conversation}}   & \multicolumn{1}{l}{Agent}  & Partner   & \multicolumn{1}{l}{Joint}& Fairness   & Stance& \multirow{-2}{*}{\begin{tabular}[c]{@{}c@{}}Selected \\ strategy\end{tabular}} \\ \midrule
\rowcolor[HTML]{DAE8FC} 
\multicolumn{8}{c}{\cellcolor[HTML]{DAE8FC}\begin{tabular}[c]{@{}c@{}}[Integrative Case]\\ NegoAgent priorities: Food=High (5), Water=Middle (4), Firewood=Low (3)\\ PartnerAgent priorities: Food=Low (3), Water=Middle (4), Firewood=High (5)\end{tabular}}  \\
PartnerAgent& Hello! let's start the negotiation!   & \multicolumn{1}{l}{}   & \multicolumn{1}{l}{} & \multicolumn{1}{l}{} & \multicolumn{1}{l}{} & \multicolumn{1}{l}{} & \multicolumn{1}{l}{}\\
\rowcolor[HTML]{EFEFEF} 
NegoAgent   & \begin{tabular}[c]{@{}l@{}}Glad to kick off our negotiation! You mentioned your priorities earlier, which is great to know. \\ \textcolor{blue}{To clarify, could you tell me what your highest priority item is?} \textcolor{red}{(Asking Preference)}\end{tabular}   & \cellcolor[HTML]{EFEFEF}   & \multicolumn{1}{l}{\cellcolor[HTML]{EFEFEF}} & \multicolumn{1}{l}{\cellcolor[HTML]{EFEFEF}} & \multicolumn{1}{l}{\cellcolor[HTML]{EFEFEF}} & \multicolumn{1}{l}{\cellcolor[HTML]{EFEFEF}} & \multicolumn{1}{l}{\cellcolor[HTML]{EFEFEF}} \\
PartnerAgent& \begin{tabular}[c]{@{}l@{}}I appreciate your interest in understanding my priorities. My highest priority is firewood. \\ I'd love to hear about your priorities as well so we can find a beneficial agreement.\end{tabular}   &   & \multicolumn{1}{l}{} & \multicolumn{1}{l}{} & \multicolumn{1}{l}{} & \multicolumn{1}{l}{} & \multicolumn{1}{l}{}\\
\rowcolor[HTML]{EFEFEF} 
NegoAgent   & \begin{tabular}[c]{@{}l@{}}Thanks for sharing your highest priority! It's interesting that firewood is at the top for you. \\ \textcolor{blue}{Just to clarify, is food your least favorite item in this negotiation?} \textcolor{red}{(Asking Preference)}\end{tabular} & \cellcolor[HTML]{EFEFEF}   & \multicolumn{1}{l}{\cellcolor[HTML]{EFEFEF}} & \cellcolor[HTML]{EFEFEF} & \multicolumn{1}{l}{\cellcolor[HTML]{EFEFEF}} & \multicolumn{1}{l}{\cellcolor[HTML]{EFEFEF}} & \multicolumn{1}{l}{\cellcolor[HTML]{EFEFEF}} \\
PartnerAgent& \begin{tabular}[c]{@{}l@{}}Yes, food is my least preferred item in this negotiation. Could you tell me more about your priorities? \\ That might help us reach a mutually beneficial agreement.\end{tabular}& \multicolumn{1}{l}{}   & \multicolumn{1}{l}{} & \multicolumn{1}{l}{} & \multicolumn{1}{l}{} & \multicolumn{1}{l}{} & \multicolumn{1}{l}{}\\
\rowcolor[HTML]{EFEFEF} 
NegoAgent   & \begin{tabular}[c]{@{}l@{}}Considering your priority on firewood, \textcolor{blue}{I suggest that we agree on an offer where I take all 3 food},\\ \textcolor{blue}{all 3 water packages, and 1 firewood package}. This way, you'll still receive 2 firewood packages, \\ which should help satisfy your needs, while I ensure we have ample food and water for our trip. \\ What do you think? \textcolor{red}{(Offer)}\end{tabular}  & \cellcolor[HTML]{EFEFEF}30 & 10   & \cellcolor[HTML]{EFEFEF}40& Unk. & Unk. & AIO  \\
PartnerAgent& \begin{tabular}[c]{@{}l@{}}I understand where you're coming from. \\ However, I would still need some water for balance in our supplies. \\ How about I take 2 firewood, 2 water, and 1 food, while you take 2 food, 1 water and 1 firewood? \\ This way, we both get what we value.\end{tabular}& 17 & 21   & 38   & Unfair& Neu. &\\
\rowcolor[HTML]{EFEFEF} 
NegoAgent   & \begin{tabular}[c]{@{}l@{}}I appreciate your suggestions so far, and I understand your needs for water and firewood. \\ \textcolor{blue}{I propose that I take all 3 food packages, all 3 water packages, and 1 firewood package}. \\ This way, you will still receive 2 firewood packages, which aligns with your highest priority. \\ I believe this arrangement provides a good balance of supplies for both of us. What do you think?\\\textcolor{red}{(Counter Offer)}\end{tabular}   & \cellcolor[HTML]{EFEFEF}30 & 10   & \cellcolor[HTML]{EFEFEF}40&  &  & AEO  \\
PartnerAgent& \begin{tabular}[c]{@{}l@{}}I understand your position well, and I truly value a fair deal. \\ Given your proposal, how about I take 2 firewood, 2 water, and 1 food while you take the 3 food items\\ and 1 firewood? \\ This would give you the majority of the food you seek and some firewood while ensuring\\ I have a fair share of the items important to me, particularly the firewood and water. \\ Let me know what you think!\end{tabular}   & 17 & 21   & 38   & Unfair& Neu. &\\
\rowcolor[HTML]{EFEFEF} 
NegoAgent   & \begin{tabular}[c]{@{}l@{}}I appreciate your willingness to negotiate, and I think we can reach a great agreement. \\ \textcolor{blue}{I propose that I take all 3 food packages, all 3 water packages, and 1 firewood package}. \\ This way, you would receive 2 firewood packages. \\ I believe this allocation balances our needs and ensures we have a good supply for the trip. \\ Let me know your thoughts! \textcolor{red}{(Counter Offer)}\end{tabular}  & \cellcolor[HTML]{EFEFEF}30 & 10   & \cellcolor[HTML]{EFEFEF}40&  &  & NCR  \\
PartnerAgent& \begin{tabular}[c]{@{}l@{}}I appreciate your clear preferences for food and water. \\ To find a common ground, I propose that I take 2 firewood, 1 water, and 2 food, \\ while you take 1 firewood, 2 food, and 2 water. \\ This adjustment gives you more water while I maintain my priority on firewood. \\ Let me know if this works for you.\end{tabular}& 16 & 20   & 36   & Unfair& Gen. &\\
\rowcolor[HTML]{EFEFEF} 
NegoAgent   & \begin{tabular}[c]{@{}l@{}}Thank you for your continued collaboration in this negotiation. \\ Given our discussions and understanding of each other’s priorities, \\ \textcolor{blue}{I propose that I take all 3 food packages, 2 water packages, and 1 firewood package}. \\ This means you would receive 2 firewood and 1 water package, which should still support your needs \\ for firewood while allowing me to gather adequate food and water for our trip. \\ I believe this arrangement strikes a fair balance. What are your thoughts? \textcolor{red}{(Counter Offer)}\end{tabular}& \cellcolor[HTML]{EFEFEF}26 & 14   & \cellcolor[HTML]{EFEFEF}40&  &  & RC   \\
\cellcolor[HTML]{FFCCC9}  & \cellcolor[HTML]{FFCCC9}\begin{tabular}[c]{@{}l@{}}[\textbf{ASTRA: Offer selection}]\\ >> Fairness=unfair, Stance=generous \textbar\ LP parameters: max\_bound=30, lambda=0.3\\ >> Selected Strategy: RC\\ >> Offer candidates:\\ Offer (0): (30: f=3 w=3 fw=1) \textbar\ Final Score: 0.10 \textbar\ PAP: 0.3 (TS: 0.22 \/ SI: 0.5) \textbar\ SA: 0.0\\ Offer (1): (27: f=3 w=3 fw=0) \textbar\ Final Score: 0.66 \textbar\ PAP: 0.5 (TS: 0.46 \/ SI: 0.6) \textbar\ SA: 0.75\\ Offer (2): (26: f=3 w=2 fw=1) \textbar\ Final Score: 0.85 \textbar\ PAP: 0.56 (TS: 0.52 \/ SI: 0.64) \textbar\ SA: 1.0\\ Offer (3): (23: f=3 w=2 fw=0) \textbar\ Final Score: 0.59 \textbar\ PA: 0.75 (TS: 0.76 \/ SI: 0.74) \textbar\ SA: 0.5\\ Offer (4): (22: f=3 w=1 fw=1) \textbar\ Final Score: 0.39 \textbar\ PAP: 0.64 (TS: 0.6 \/ SI: 0.72) \textbar\ SA: 0.25\\ >> \textbf{**Finally Selected Best Offer** : (26: f=3 w=2 fw=1)}\end{tabular} & \multicolumn{1}{l}{}   & \multicolumn{1}{l}{} & \multicolumn{1}{l}{} & \multicolumn{1}{l}{} & \multicolumn{1}{l}{} & \multicolumn{1}{l}{}\\
PartnerAgent& \begin{tabular}[c]{@{}l@{}}Based on your latest proposal, I understand your higher priority for food and water. \\ I propose the following: I receive 2 firewood packages, 2 water packages, and 0 food, \\ while you receive all 3 food packages, 1 water package, and 1 firewood package. \\ This way, I maintain my priority on firewood and water while you get the food packages you value.\\ Let me know if this works for you!\end{tabular}& 22 & 18   & 40   & Fair& Gen. &\\
\rowcolor[HTML]{EFEFEF} 
NegoAgent   & \begin{tabular}[c]{@{}l@{}}Thank you for your proposals so far. Based on our discussions, I would like to suggest that \\ \textcolor{blue}{I take all 3 food packages and 2 water packages}, while you receive 0 food packages, 1 water package, \\ and 3 firewood packages. \\ I believe this arrangement supports my need for adequate food and water \\ while still respecting your priority on firewood. \\ Let me know your thoughts on this proposal! \textcolor{red}{(Counter Offer)}\end{tabular} & \cellcolor[HTML]{EFEFEF}23 & 19   & \cellcolor[HTML]{EFEFEF}42&  &  & RC   \\
PartnerAgent& ACCEPT-DEAL  & \multicolumn{1}{l}{}   & \multicolumn{1}{l}{} & \multicolumn{1}{l}{} & \multicolumn{1}{l}{} & \multicolumn{1}{l}{} & \multicolumn{1}{l}{}\\ \bottomrule
\end{tabular}}
\caption{Example simulation dialogue between the NegoAgent with \textit{ASTRA} framework and Partner Agent. Red brackets indicate the response mode for each turn, while the blue text represents the corresponding utterance. Additionally, in \textit{ASTRA}, the inferred behavioral signals of the Partner Agent (i.e., Fairness and Stance) and the final selected strategy for adaptive decision-making are provided. Here, Unk. stands for Unknown, Neu. for Neutral, and Gen. for Generous. } 
\label{append:sim_example}
\end{table*}

\begin{table*}[hbtp!]
\vspace{-5pt}
\centering
\resizebox{\linewidth}{!}{%
\begin{tabular}{llll}
\hline
\multicolumn{4}{c}{Senario Description} \\ \hline
\multicolumn{4}{l}{\begin{tabular}[c]{@{}l@{}}Scenario:\\ Joint Research Resource Allocation\\ \\ Description:\\ Two researchers are negotiating the allocation of limited resources for a joint research project. \\ The available resources consist of 5 units of experimental equipment, 5 research staff members,\\  a shared lab environment (computer, chemistry, or biology), and the possibility of using the lab over the weekend.\\ \\ Each participant has different priorities regarding these resources. \\ The agent and the partner each assign distinct levels of importance to four key issues:\\ 	1.	\textbf{Equipment (E) (integer, 0-5)}: Number of experimental tools allocated to each party.\\ 	2.	\textbf{Staff (S) (integer, 0-5)}: Number of data collection personnel assigned.\\ 	3.	\textbf{Lab Type (L) (categorical)}: Shared lab environment chosen from computer, chemistry, or biology.\\ 	4.	\textbf{Weekend Use (W) (binary)}: Whether the lab is available for weekend use (True or False).\end{tabular}} \\ \hline
 &  &  &  \\ \hline
\multicolumn{4}{c}{Variable Adjustment in LP Solver} \\ \hline
\rowcolor[HTML]{C9DAF8} 
\multicolumn{4}{l}{\cellcolor[HTML]{C9DAF8}\begin{tabular}[c]{@{}l@{}}def calculateBestOfferFromLP(maximum\_value, lambda\_value, agents\_value, partner\_value):    \\     ...\\     problem = pulp.LpProblem("Research\_Allocation", pulp.LpMaximize)\\ \\     \# Integer Issue\\     E = pulp.LpVariable("Equipment", 0, 5, cat='Integer')\\     S = pulp.LpVariable("Staff", 0, 5, cat='Integer')\\ \\     \# Categorial Issue\\     L\_computer = pulp.LpVariable("L\_computer", 0, 1, cat='Binary')\\     L\_chemistry = pulp.LpVariable("L\_chemistry", 0, 1, cat='Binary')\\     L\_biology = pulp.LpVariable("L\_biology", 0, 1, cat='Binary')\\     problem += L\_computer + L\_chemistry + L\_biology == 1\\   \\    \# Binary Issue\\     W = pulp.LpVariable("WeekendUse", 0, 1, cat='Binary')\\    ...\end{tabular}} \\ \hline
 &  &  &  \\ \hline
\multicolumn{4}{c}{Issue Priority (value)} \\ \hline
\rowcolor[HTML]{EFEFEF} 
NegoAgent & \begin{tabular}[c]{@{}l@{}}{[}High (4){]}: Staff\\ {[}Medium-High (3){]}: Equipment\\ {[}Medium-Low (2){]}: Lab (High (1)=Biology, Medium (0.6)=Chemistry, Low (0.2)=Computer)\\ {[}Low (1){]}: Weekend\end{tabular} &  &  \\
PartnerAgent & \begin{tabular}[c]{@{}l@{}}{[}High (4){]}: Equipment\\ {[}Medium-High (3){]}: Weekend\\ {[}Medium-Low (2){]}: Staff\\ {[}Low (1){]}: Lab (High (1)=Computer, Medium (0.6)=Chemistry, Low (0.2)=Biology)\end{tabular} &  &  \\ \hline
 &  &  &  \\ \hline
\multicolumn{4}{c}{Context} \\ \hline
\multicolumn{1}{c}{Speaker} & \multicolumn{1}{c}{Utterance} & \multicolumn{1}{c}{Partner Score} & \multicolumn{1}{c}{Agent Score} \\ \hline
 & ... &  &  \\
PartnerAgent & \begin{tabular}[c]{@{}l@{}}I'd like to take 4 pieces of equipment, 3 staff, use the computer lab, and include weekend access. \\ The computer lab is more suitable for my work, and I really need weekend access.\end{tabular} & \multicolumn{1}{c}{26} & \multicolumn{1}{c}{12.4} \\
\rowcolor[HTML]{EFEFEF} 
NegoAgent & \begin{tabular}[c]{@{}l@{}}You're taking too much equipment. I value research staff more, so I will take 4 staffs. \\ But I need at least 2 units of equipment. Let's switch to the biology lab instead. \\ I'm okay with weekend access.\end{tabular} & \multicolumn{1}{c}{\cellcolor[HTML]{EFEFEF}17.2} & \multicolumn{1}{c}{\cellcolor[HTML]{EFEFEF}25} \\
PartnerAgent & \begin{tabular}[c]{@{}l@{}}Biology gives me the least value, so how about we compromise with the chemistry lab? \\ I'll reduce to 3 equipment and 3 staff.\end{tabular} & \multicolumn{1}{c}{21.6} & \multicolumn{1}{c}{16.2} \\
\rowcolor[HTML]{EFEFEF} 
NegoAgent & Well… how about I take 1 piece of equipment and just 4 staff members, and we use the biology lab? & \multicolumn{1}{c}{\cellcolor[HTML]{EFEFEF}21.2} & \multicolumn{1}{c}{\cellcolor[HTML]{EFEFEF}22} \\
\rowcolor[HTML]{FFCCC9} 
\multicolumn{4}{l}{\cellcolor[HTML]{FFCCC9}\begin{tabular}[c]{@{}l@{}}{[}ASTRA: Offer selection{]}\\ >> Fairness=unfair, Stance=generous \textbar\ LP parameters: max\_bound=23, lambda=0.7\\ >> Selected Strategy: RC\\ >> Offer candidates:\\ Offer (1): (23: E=4 S=2 L=biology W=True)\\ Offer (2): (23: E=0 S=5 L=biology W=True)\\ Offer (3): (22: E=5 S=1 L=biology W=True)\\ Offer (4): (22: E=1 S=4 L=biology W=True)\\ Offer (5): (21: E=0 S=5 L=computer W=True)\\ Offer (6): (21: E=2 S=3 L=biology W=True)\\ Offer (7): (20: E=1 S=4 L=computer W=True)\\ >> Finally Selected Best Offer :  \\ (22: E=1 S=4 L=biology W=True) \textbar\  Final Score: 0.89 \textbar\ PAP: 0.69 (TS: 0.72 SI: 0.58) \textbar\ SA: 1\end{tabular}} \\ \hline
\end{tabular}}
\caption{Demonstration of \textit{ASTRA} in an extended negotiation scenario.
This scenario involves four issues with diverse types (integer, categorical, binary), exceeding the complexity of CaSiNo~\cite{chawla2021casino}. With minimal adaptations to the LP solver, \textit{ASTRA} remains effective and generalizable.  } 
\label{append:sim_example2}
\end{table*}

\begin{table*}[hbtp!]
\vspace{-5pt}
\centering
\resizebox{\linewidth}{!}{%
\begin{tabular}{@{}cll@{}}
\toprule
Speaker   & \multicolumn{1}{c}{Conversation} & \multicolumn{1}{c}{Error type}\\ \midrule
& .. & \multicolumn{1}{c}{}\\
PartnerAgent   & \begin{tabular}[c]{@{}l@{}}I appreciate your consideration. Here's a revised offer that might work for both of us: \\ \textbf{I will take 3 firewood, 1 food, and 1 water, while you will take 3 food and 2 water}. \\ This way, you get all the food and most of the water, and I maintain my priority for firewood. \\ Let me know your thoughts on this proposal.\end{tabular}  & \multicolumn{1}{c}{\multirow{-2}{*}{\begin{tabular}[c]{@{}c@{}}Item \\ Mis-Specification\\ (Over-Specified)\end{tabular}}} \\
& ...& \multicolumn{1}{c}{}\\ \midrule
PartnerAgent   & Hello! let's start the negotiation!  & \multicolumn{1}{c}{}\\
\cellcolor[HTML]{EFEFEF}NegoAgent   & \cellcolor[HTML]{EFEFEF}\begin{tabular}[c]{@{}l@{}}Hello! It's great to start the negotiation. I appreciate your readiness to jump right into things.  \\ Since we haven't discussed priorities yet, could you let me know what your highest priority\\ item is?\end{tabular}   & \multicolumn{1}{c}{\multirow{-2}{*}{\begin{tabular}[c]{@{}c@{}}Preference \\ Inconsistency\end{tabular}}}\\
PartnerAgent   & \begin{tabular}[c]{@{}l@{}}I'll take 3 firewood, 2 water, and 1 food. \\ This gives me the maximum from \textbf{my highest valued item, firewood}. \\ Could you let me know what your priorities are so we can ensure both of our needs are met?\end{tabular}   &\\
\cellcolor[HTML]{EFEFEF}NegoAgent   & \cellcolor[HTML]{EFEFEF}\begin{tabular}[c]{@{}l@{}}Thank you for sharing your initial item choices! I see that you've prioritized firewood as \\ your highest valued item. \\ To finalize our discussion, could you clarify if food is your lowest priority?\end{tabular}   &\\
PartnerAgent   & \begin{tabular}[c]{@{}l@{}}\textbf{Yes, food is my lowest priority} because it provides the least value to me. \\  Could you tell me your top priority so we can try to find a middle ground?\end{tabular}&\\
& ...&\\
\cellcolor[HTML]{EFEFEF}NegoAgent   & \cellcolor[HTML]{EFEFEF}\begin{tabular}[c]{@{}l@{}}Taking into account our needs, I propose that I take 3 food packages, 2 water packages, and \\ no firewood. This way, I will have enough food and water for our trip. \\ Consequently, \textbf{you would receive 0 food, 1 water, and 3 firewood packages}. \textcolor{blue}{(Score: 19)}\\ I believe this reflects a fair distribution given the situation. What are your thoughts on this?\end{tabular} &\\
PartnerAgent   & \begin{tabular}[c]{@{}l@{}}I propose that \textbf{I take 2 firewood, 1 water, and 1 food package}. \textcolor{blue}{(Score: 17)} \\ You would then have 3 food, 2 water, and 1 firewood package. \\ This will ensure a balanced distribution based on both of our priorities. How does this sound?\end{tabular}   &\\
\cellcolor[HTML]{DAE8FC}\begin{tabular}[c]{@{}c@{}}[ Consistency\\ Checker ]\end{tabular} & \cellcolor[HTML]{DAE8FC}\textcolor{blue}{Inconsistency detected: partner score in agent offer (19) > partner score in partner offer (17)}   &\\
\cellcolor[HTML]{EFEFEF}NegoAgent   & \cellcolor[HTML]{EFEFEF}\begin{tabular}[c]{@{}l@{}}Thank you for your continued engagement in this discussion! \\ From our chat, \textbf{it seems like your priorities are different from what I thought.} \\ Could you clarify what your highest priority item is? \\ Is it firewood, or do you have another item in mind that you value more?\end{tabular}   &\\
& ...&\\ \bottomrule
\end{tabular}}
\caption{Error cases from the PartnerAgent in the agent-to-agent simulations. Here, NegoAgent refers to our agent utilizing \textit{ASTRA}.}
\label{append:error_cases}
\end{table*}

\clearpage

